\documentclass[10pt,twocolumn,letterpaper]{article}

\usepackage{cvm}
\usepackage{times}
\usepackage{graphicx}
\usepackage{amsmath,amssymb}
\usepackage{booktabs}
\usepackage[table]{xcolor}
\usepackage{multirow}
\usepackage{framed}
\usepackage{CJKutf8}
\usepackage{lineno}
\usepackage{algorithm,algpseudocode}
\usepackage{threeparttable}
\usepackage{subcaption}
\usepackage[numbers]{natbib}
\usepackage{tabularx}
\usepackage{setspace}
\usepackage{placeins}
\usepackage[breaklinks=true,letterpaper=true,colorlinks,bookmarks=false]{hyperref}
\usepackage[capitalize]{cleveref}

\cvmfinalcopy
\ifcvmfinal\fi
\newcommand{\keywords}[1]{{\bf\emph{Keywords: #1}}}

\newcommand{\ZY}[1]{{\color{black}#1}}
\newcommand{\ZZ}[1]{{\color{black}#1}}
\newcommand{\ZFL}[1]{{\color{black}#1}}
\newcommand{\LT}[1]{{\color{black}#1}}

\begin{document}

\title{AttenIR: Aligning Dense Visual Attention with Mesh Warping for Unsupervised Image Retargeting}
\author{%
{\small Yun Zhang$^{1}$ \enspace Zhiruo Zhang$^{1}$\thanks{Corresponding author, \texttt{zrzhang0101011@gmail.com}} \enspace Lifei Zhu$^{1}$ \enspace Zhe Zhu$^{2}$ \enspace Tong Li$^{1}$ \enspace Fang-Lue Zhang$^{3}$ \enspace Shengjuan Liu$^{1}$ \enspace Xinyuan Zheng$^{1}$}\\[2pt]
{\footnotesize $^{1}$School of Media Engineering, Communication University of Zhejiang, Hangzhou 310018, China}\\
{\footnotesize $^{2}$Samsung Research America, Irvine, CA 92612, USA}\\
{\footnotesize $^{3}$School of Computer Science and Engineering, The University of New South Wales, Sydney, NSW 2033, Australia}%
}
\maketitle

\begin{abstract}
Image retargeting aims to resize images while preserving salient content free from distortion and stretching.
Recent learning-based solutions reduce \ZZ{reliance} on traditional hand-crafted features and \ZZ{optimization procedures}, \ZZ{leading to} more efficient and generalizable performance.
Nevertheless, \ZZ{existing} approaches still \ZZ{struggle to preserve global structure and content fidelity}. 
To address these \ZZ{limitations}, we propose \LT{\textit{AttenIR},} an unsupervised image retargeting \ZZ{framework} requiring no manual labels, based on content-aware mesh warping, which effectively couples dense visual attention with sparse mesh constraints.
We formulate the mesh deformation as a \ZFL{vertex displacement} regression task. 
Guided by saliency priors, \ZZ{\ZFL{our} method jointly optimizes scale, structure, and geometry preservation losses. \ZFL{By combining dense visual attention with semantic features, it achieves robust, unsupervised retargeting without manual annotations.}}
Our primary contributions include (1) a lightweight unsupervised method for \ZFL{image} retargeting that ensures content fidelity and preserves global structure, and (2) a coarse-to-fine projection strategy to align dense visual attention priors with sparse mesh constraints, enabling content-aware deformation at the pixel level.
Experiments demonstrate that our method achieves competitive \LT{and} superior performance compared with \ZZ{both} traditional and state-of-the-art learning-based methods.
\end{abstract}


 %\linenumbers


\section{Introduction}
\ZZ{\ZFL{Image retargeting adjusts aspect ratios while preserving salient content and minimizing structural distortion. As visual media is increasingly viewed across diverse screen sizes, retargeting has become a crucial problem in computer graphics and image processing.}}
%Image retargeting adjusts image resolution or aspect ratio while keeping the main content complete and minimizing structural distortion, enabling high-quality display across devices with diverse screen specifications. With the rapid popularization of smartphones, tablets, and other portable devices, visual content is increasingly viewed under heterogeneous display conditions, making image retargeting an important problem in computer graphics and image processing.}
% Image retargeting aims to adjust the resolution and aspect ratio of images. 
% It keeps the main content complete and avoids structural distortion, enabling high-quality visual display across various devices with different screen specifications.
% With the rapid popularization of smartphones, tablets, and other portable devices, diverse visual display scenarios have emerged, and consequently, making image retargeting a prominent research focus in the fields of computer graphics and image editing.
\ZZ{Image retargeting methods can be broadly divided into traditional optimization-based approaches and learning-based approaches. Traditional methods perform content-aware resizing with operators such as seam carving~\cite{avidan2007seam}, mesh-based warping~\cite{guo2009mesh,kim_quad_2019,cao2023polygonal}, and operator-based editing~\cite{simakov2008summarizing,pritch2009shift}. To reduce reliance on hand-crafted features and iterative optimization, learning-based methods have attracted increasing attention because they can capture semantic representations and provide stronger robustness and generalization. 
Representative works~\cite{cho2017weakly,danon2021deepfeatures,le2024retargeting,liao2026object} benefit from combining deep semantic features with content-aware geometric warping. 
However, most warping-based methods struggle to bridge the visual attention prior and mesh warping, leading to undesirable structure and content distortion.
\ZFL{To overcome the limitations of warping-based methods and improve visual quality, recent approaches adopt generative models to} retarget images 
via content removal, completion, or relocation~\cite{shen2024prune,dong2026airdr}. 
\ZFL{While effective in some scenarios, generative models are prone to content hallucination during retargeting, failing to faithfully preserve the original input details.}}
% Generally, image retargeting can be classified into two main categories: traditional optimization-based methods and learning-based methods.
% Traditional methods are designed to retarget images in a content-aware manner using seam-carving~\cite{avidan2007seam}, mesh-based warping~\cite{guo2009mesh,kim_quad_2019,cao2023polygonal}, operator-based approaches~\cite{simakov2008summarizing,pritch2009shift}, etc.
% To reduce the reliance on handcrafted features and optimizations, learning-based methods stand out, as they can effectively capture semantic features to achieve stronger robustness and generalization capability.
% Representative works include~\cite{cho2017weakly,danon2021deepfeatures,le2024retargeting,liao2026object}, etc, which benefit from combining deep semantic features with content-aware geometric warping.
% To avoid the severe distortions introduced by warping-based solutions and to produce more visually pleasing results, generative models are adopted to retarget images by content removal, completion, etc~\cite{shen2024prune,dong2026airdr}.
% Although effective in various scenarios, generative models tend to produce novel content during retargeting, which not only struggles to faithfully retain original input information, but also tends to produce unnatural and implausible visual artifacts.

To address the aforementioned limitations, 
we propose \LT{\textit{AttenIR},} an unsupervised retargeting framework requiring no manual labels, which aligns dense visual attention with mesh constraints to mitigate geometric warping distortions and preserve the fidelity of the original content.
Specifically, ``\textit{dense visual attention}'' denotes a pixel-level saliency prior~\cite{ChengPAMI,SalObjBenchmark} employed to guide mesh warping, which is obtained from a pretrained saliency model~\cite{gao2024mdsam}.
% and we adopt saliency prior to characterize the visual attention of input images.
% we propose a lightweight unsupervised retargeting framework to alleviate geometric warping distortions \ZZ{while maintaining the fidelity of the original content.}  
In our method, we adopt mesh-based warping to adapt image geometry for target aspect ratios, where mesh deformation is formulated as a \ZZ{vertex displacement} regression task. 
Guided by saliency priors, the regression is constrained by content-preserving, global structure-preserving, and local shape-preserving loss terms to \ZZ{maintain} visual plausibility.
% To effectively align the dense visual attention data with sparse mesh constraints, 
% we propose a \ZZ{fine-grained warping scheme to align mesh deformation with dense saliency maps. 
Instead of averaging dense saliency values into coarse mesh-cell weights, we project mesh-induced deviations derived from the loss functions to the pixel level. 
\ZFL{Specifically, the saliency prior is warped using the deformed mesh, aligning visual attention and geometric displacement within a shared output space. This enhances salient structure preservation while maintaining} the flexibility of mesh-based warping. Experiments demonstrate that our unsupervised method achieves robust retargeting and generalizes well across diverse scenarios.

In summary, the main contributions of this paper are as follows: 
\begin{itemize}
    \item \ZZ{We propose a lightweight unsupervised \ZFL{learning framework} for image retargeting that preserves content fidelity and global structure without requiring manually labeled training data.}
    \item \ZZ{We develop a coarse-to-fine projection strategy to align dense visual attention priors with sparse mesh constraints, enabling content-aware deformation at the pixel-level.}
\end{itemize}

\section{Related Work}
\subsection{Classical Content-Aware Retargeting}
Image retargeting adapts an image to a target resolution or aspect ratio while preserving the perceptually important content and maintaining plausible geometry.
Unlike uniform scaling and cropping, which respectively distort the whole image or discard content without considering visual importance, content-aware methods explicitly estimate where resizing operations can be applied with lower perceptual cost.
Early discrete methods are represented by seam carving~\cite{avidan2007seam}, which removes or inserts connected low-energy seams and thereby changes image size with a simple and effective local operator.
Forward-energy seam carving and its video extension further reduce artifacts introduced by seam removal~\cite{rubinstein2008improved}, while optimized seam carving combined with scaling balances pixel removal and global deformation~\cite{dong2009optimized}.
Since no single operator is reliable for all images and all target sizes, multi-operator retargeting combines cropping, scaling, and seam carving through an optimized operator sequence~\cite{rubinstein2009multi}.
Patch- and correspondence-based formulations, such as bidirectional similarity~\cite{simakov2008summarizing} and shift-map editing~\cite{pritch2009shift}, broaden the search space by rearranging image content rather than only deleting seams.
These operator-based approaches are flexible and intuitive, but they may break long structures, remove thin objects, or introduce local discontinuities when salient objects and background textures are strongly interleaved.

\subsection{Warping and Mesh-Based Deformation}
A complementary line formulates retargeting as a continuous deformation problem.
Wolf et al.~\cite{wolf2007non} introduced non-homogeneous content-driven video retargeting, and optimized scale-and-stretch~\cite{wang2008optimized} distributes resizing distortion over an image grid under saliency-aware constraints.
Subsequent methods improve geometric plausibility by preserving shapes~\cite{zhang2009shape}, line structures~\cite{chang2012line}, and axis-aligned deformation behavior~\cite{panozzo2012robust}.
Mesh-based methods are especially relevant to our work because they provide an explicit and controllable deformation space.
Guo et al.~\cite{guo2009mesh} incorporate saliency and structure into a mesh parametrization, while subsequent grid- or polygonal finite-element formulations improve the representation of local image geometry and deformation constraints~\cite{kim_quad_2019,cao2023polygonal}.
Recent deformation-field retargeting further generalizes content-aware deformation beyond 2D images and learns where deformation can be applied with low information loss~\cite{elsner2024deformation}.
Compared with discrete deletion or recomposition, deformation-based retargeting preserves a direct source-to-target correspondence, which is desirable when the retargeted results must remain faithful to the original scene.
However, conventional deformation methods often rely on hand-designed saliency maps and cell-level objectives; compressing dense visual attention into a single mesh-cell weight can lose small salient regions and fine object boundaries.
This limitation motivates a tighter alignment between pixel-level visual attention and mesh-level geometric regularization.

\subsection{Learning-Based and Generative Retargeting}
Deep learning methods reduce the dependence on manually designed energy functions by learning semantic representations, deformation policies, or feature-space resizing objectives.
Cho et al.~\cite{cho2017weakly} proposed a weakly and self-supervised CNN that predicts shift maps while implicitly learning attention from image-level supervision.
Deep multi-operator retargeting learns to select resizing operators for photo squarization~\cite{song2018photo}, while DeepIR~\cite{lin2019deepir} uses semantic features in a coarse-to-fine framework.
Cycle-IR~\cite{tan2019cycle} introduces cyclic perceptual consistency so that a retargeted image can be mapped back to the source with limited semantic loss, and feature-reconstruction methods resize images by manipulating deep feature maps rather than raw pixels~\cite{danon2021deepfeatures}.
Learning-based retargeting has also been extended to video through end-to-end adaptive deformation~\cite{le2024retargeting}.
More recent work revisits the connection between learning and explicit geometry: Object-IR~\cite{liao2026object} casts image retargeting as self-supervised mesh warping guided by object consistency and geometric preservation, and HALO~\cite{xu2026halo} decomposes an image into salient and non-salient layers that are warped with different fields.
These methods show the benefit of combining semantic awareness with geometric deformation, but they still leave open how dense saliency should be coupled to mesh constraints across scales.

Generative models provide another direction by treating retargeting as content removal, relocation, and image completion.
PruneRepaint~\cite{shen2024prune} uses semantic importance to prune less important regions and repaint local content for arbitrary aspect ratios, while AIR-DR~\cite{dong2026airdr} relocates foreground instances and uses dual-guided repainting to recover background regions.
Such methods are effective for aggressive aspect-ratio changes and can produce visually plausible images, but they may synthesize or relocate content beyond the original pixel correspondence.
For applications where the output should remain a faithful geometric transformation of the input, a deformation-based formulation remains preferable.

\subsection{Visual Attention and Retargeting Quality}
Across classical, learning-based, and generative retargeting, a central question is how to represent visual importance.
Classical methods use gradients, edges, faces, or saliency maps to suppress distortion in important regions~\cite{avidan2007seam,wang2008optimized,guo2009mesh}.
Semantic segmentation and pixel-fusion strategies introduce stronger object-level priors~\cite{niu2020semantic}, and recent foundation-model-based salient object detectors such as MDSAM~\cite{gao2024mdsam} provide more detailed masks with stronger generalization.
These advances make it possible to guide retargeting with dense visual attention, but they also expose a scale mismatch: visual attention is naturally defined at the pixel or object-boundary level, whereas mesh deformation is controlled by sparse vertices and cells.
A coarse cell-wise aggregation of saliency is therefore insufficient for preserving small objects and fine structures.

Evaluation studies reinforce the importance of semantic and geometric preservation.
The RetargetMe benchmark and comparative user study established a perceptual basis for comparing representative retargeting operators~\cite{rubinstein2010comparative}.
Objective retargeting quality assessment further models content loss and shape distortion~\cite{liu2011quality,ma2012quality}, and recent surveys summarize the development of retargeting algorithms, datasets, and quality metrics~\cite{guo2023irqa,fan2024review}.
These studies indicate that retargeting quality is not determined only by global image similarity; users are particularly sensitive to deformation, structure breakage, and content loss in salient regions.
Therefore, our work follows the faithful deformation-based line but introduces the content-aware warping based on the alignment of dense visual attention with sparse mesh constraints, preserving important structures while allowing less salient regions to absorb necessary aspect-ratio changes.

\section{Method}
\begin{figure*}[!t]
	\centering
	\includegraphics[width=\linewidth]{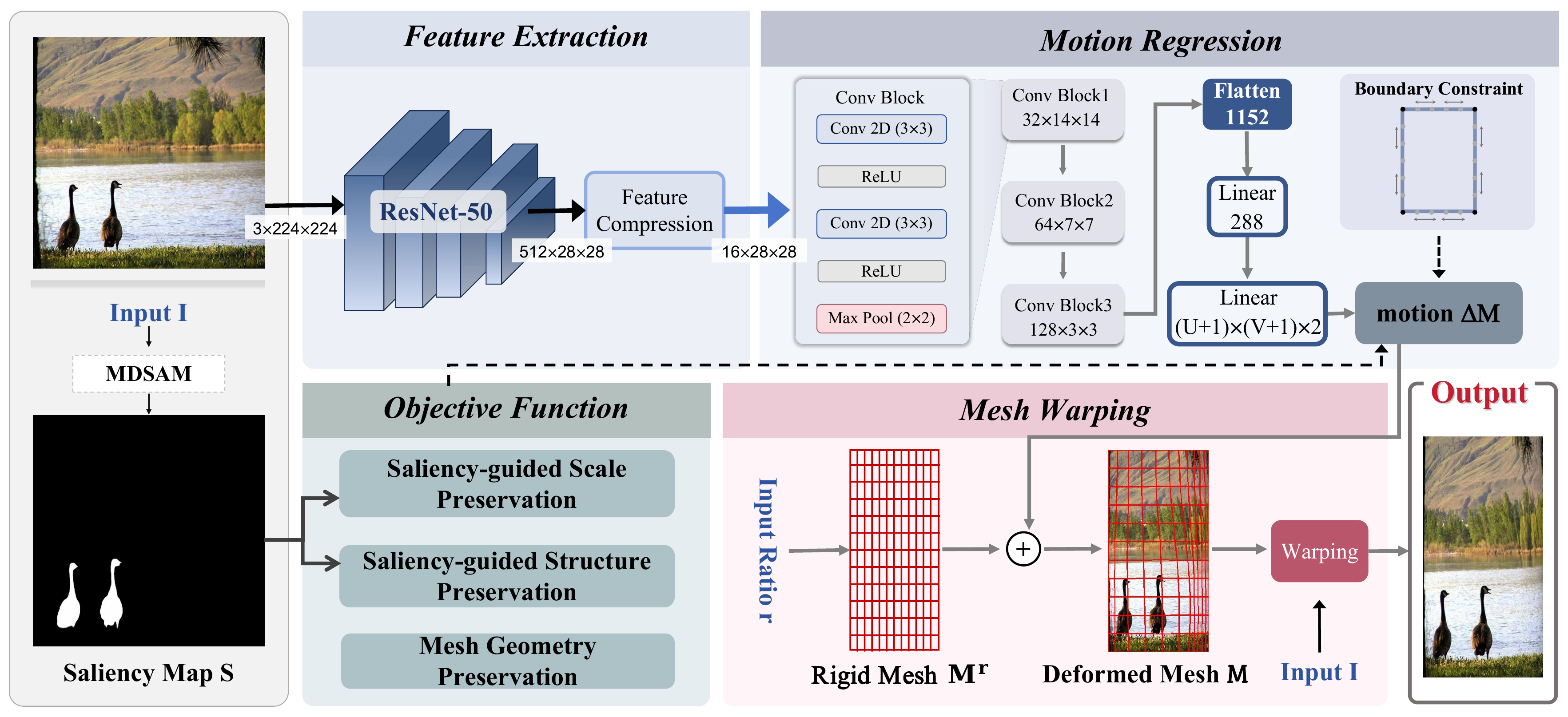}
	\caption{\ZZ{Overall network architecture of the proposed method.
    The network takes an image and a target aspect ratio as input and outputs the \ZFL{vertex displacement} of the original rigid mesh, which is then used to warp the input image and produce the retargeting result.}}
	\label{framework}
\end{figure*}

\ZY{As shown in Fig.~\ref{framework}, our method takes the original image and the target aspect ratio as inputs and outputs a content-aware retargeted image \ZZ{through \ZFL{mesh warping learned in an unsupervised manner}. The framework is constrained by a comprehensive objective that encourages uniform scaling of salient regions while preserving global structures and local mesh geometry. }
 % constrained by a comprehensive objective functions, which ensure the uniform scaling of the salient regions, the preservation of global structures and local shapes.
To avoid the degraded retargeting caused by coarse saliency averaged over mesh cells, we propose an unsupervised retargeting framework that effectively aligns dense visual attention with mesh constraints.
}
% Our network is optimized by a comprehensive objective function composed of three losses: 
% the saliency-aware content-preserving loss $L_c$, the global structure-preserving loss $L_g$, and the local shape-preserving loss $L_s$. 
% Specifically, $L_c$ \ZY{is employed to enforce uniform scaling of the salient regions so as to keep important content geometrically undistorted;} $L_g$ regularizes the distribution of geometric deviations \ZY{across} the entire mesh to maintain global structural consistency.
% Here, geometric deviation refers to the distortion induced by mesh deformation during image resizing. 
% In contrast, $L_s$ serves as a geometric validity constraint that prevents undesirable local configurations such as mesh folding and cell collapse. Based on the geometric deviations characterized by $L_c$ and $L_g$, we further introduce a multiscale visual attention alignment mechanism. 
% Instead of averaging dense saliency values into coarse mesh-cell weights, the proposed mechanism projects mesh-induced geometric deviations into the output pixel space and warps the saliency map with the same deformed mesh. Building Pixel-level Fine-Grained Constraints}}

\ZFL{In the following sections, we describe our unsupervised image retargeting \ZFL{framework} in Sec.~\ref{network}, introduce fine-grained pixel-level constraints in Sec.~\ref{finegrain}, and detail the objective functions in Sec.~\ref{loss}.}

\ZY{\subsection{Unsupervised Image Retargeting via Warping}\label{network}}
% \subsection{Mesh-Based Retargeting Backbone}

To provide a \ZY{light-weighted} and \ZY{effective solution} 
% controllable and differentiable deformation space 
for image retargeting, we adopt \ZY{the traditional} mesh warping model~\cite{liao2026object}, where retargeting is achieved by predicting vertex displacements on a rigid quad mesh.
Given an input image $I \in \mathbb{R}^{H \times W \times 3}$ and a target aspect ratio $r$, we first construct a uniform rigid mesh $M^r$ over the image \ZY{with a resolution of $U \times V$, which consists of $(U+1)\times(V+1)$ vertices.
Then, a convolutional regression network is constructed to predict the}
vertex-wise \ZFL{displacement} field $\Delta M$, and the deformed mesh is obtained by
\begin{equation}
\mathcal{M} = M^r + \Delta M.
\end{equation}

% \ZY{Specifically,}
% For a mesh resolution of $U \times V$, the mesh contains $(U+1)\times(V+1)$ vertices, denoted as 
% $\{v_{a,b}\mid a\in[0,U], b\in[0,V]\}$, where $a$ and $b$ indicate the row and column indices of a mesh vertex, respectively.
% Accordingly, the regression output has a dimension of $(U+1)\times(V+1)\times2$.
% Specifically, we denote the motion of vertex $v_{a,b}$ as 
% $\Delta M_{a,b}=(\Delta x_{a,b},\Delta y_{a,b})$, where $a\in[0,U]$ and $b\in[0,V]$.

\ZY{Finally}, the retargeted image $\mathcal{J}$ is \ZY{generated} by warping the input image according to the correspondence between the deformed mesh and the target rigid mesh, \ZY{$\mathcal{J} = \mathcal{W}(I, \mathcal{M})$,
where $\mathcal{W}$ refers the Thin-plate spline transformation (TPS)~\cite{TPS_Transformation} that provides more natural and smoother transitions.}
% This formulation keeps the retargeting process explicitly geometric: the network does not synthesize pixels directly, but predicts the deformation of mesh vertices.

\ZY{Similar to previous learning-based image warping tasks~\cite{ZHANG2026104615,liao2026object}}, 
we also employ ResNet-50~\cite{DBLP:conf/cvpr/HeZRS16} feature extractor with pre-trained parameters to obtain dense semantic representations of the input image. 
\ZZ{Instead of} directly feeding the extracted high-dimensional feature map into the regression head, we introduce a compact feature projection module before \ZY{the} \ZFL{vertex} regression. 
\ZFL{This module projects features into a compact, deformation-aware representation to reduce redundancy and stabilize the input for the subsequent vertex displacement regression\ZY{, which }predicts the final vertex offsets.}
%This module maps the extracted feature map into a lower-dimensional deformation-oriented representation, thereby reducing feature redundancy and providing a more stable input for the subsequent \ZFL{vertex displacement} regression\ZY{, which }estimates the displacement of each mesh vertex. 

To stabilize the deformation and avoid irregular output boundaries, we further apply a boundary constraint to the motion field. 
\ZFL{Top and bottom boundary vertices are restricted to horizontal sliding, while left and right boundary vertices are constrained to vertical sliding.
}
The four corner vertices are fixed. 
This design prevents boundary drift while preserving sufficient tangential flexibility for aspect-ratio adaptation.

\ZY{As no ground-truth labels are accessible for image retargeting tasks, our displacement regression is optimized via an unsupervised framework built upon multiple complementary loss functions: }
the saliency-guided scale preservation loss $\mathcal{L}_{c}$, the saliency-guided structure preservation loss $\mathcal{L}_{t}$, and the mesh geometry preservation loss $\mathcal{L}_{g}$. 
\ZY{To fully leverage dense saliency for constraining scale variation of salient regions, we incorporate a pixel-level fine-grained constraint to couple dense visual attention with cell-level mesh constraints.}
% Based on this mesh-based deformation backbone, the remaining problem is to align visual attention with mesh-level deformation, enabling the deformed mesh to preserve important structures while absorbing distortion in less-attended regions.

\subsection{\ZY{Aligning Dense Saliency with Sparse Mesh Constraints}}
\label{finegrain}

Visual attention~\cite{ChengPAMI,SalObjBenchmark} has been widely used as an effective prior in content-aware image retargeting, such as~\cite{kim_quad_2019,wang2008optimized}.
\ZY{Inspired by these methods, we also focus on preserving \ZZ{the scale and structure of visually salient regions within} our retargeting framework.}

\begin{figure*}[!t]
	\centering	\includegraphics[width=\linewidth]{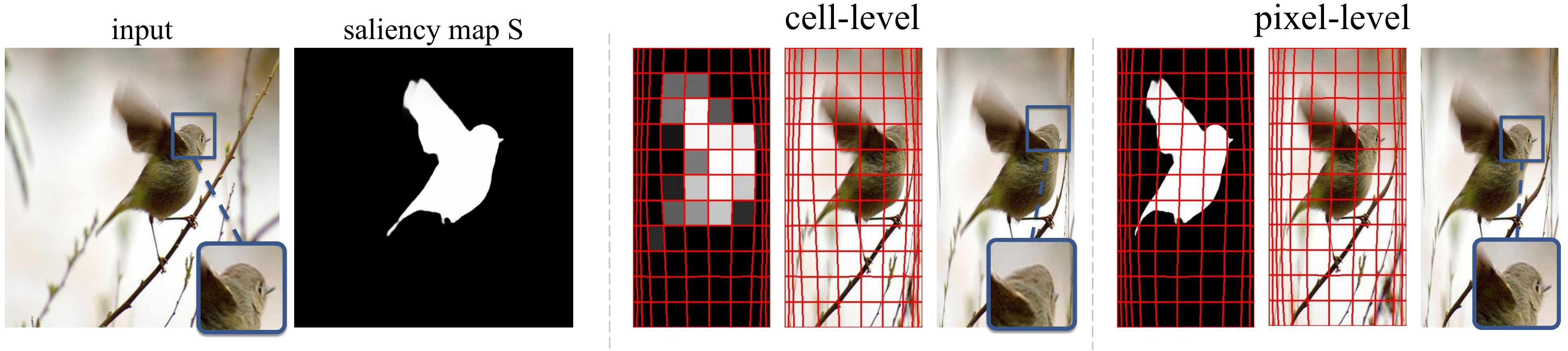}
	\caption{Comparison \ZY{\ZZ{between} results generated by cell-level coarse constraints and pixel-level fine-grained constraints. 
    The zoom-in views indicate that methods employing cell-level constraints, obtained by averaging dense saliency values over each mesh cell, cannot adequately preserve appearance and structural features.
    In contrast, our pixel-level constraints yield more favorable results.}
    % Two examples show that the bird beak and the butterfly wing edge, as representative small regions and fine boundaries, are easily diluted or blurred at the cell level.
    % Zoom in for the best view.
}
    \label{pix_cell}
\end{figure*}

% consider visually attractive and salient image regions as the key important regions in our method.
\ZY{
However, it is nontrivial to build effective connections between dense saliency and sparse \ZZ{mesh deformation}. 
As shown in Fig.~\ref{pix_cell}, cell-level coarse constraints, which are generated by averaging pixel-level dense saliency across each mesh cell to obtain cell-wise saliency, fail to fully preserve salient structures in warping-based retargeting. 
This limitation primarily stems from the inherent difficulty in aligning sparse mesh-level constraints with dense saliency priors. 
To address this challenge, \ZFL{We introduce pixel-wise constraints to precisely align the warped image with dense saliency priors while preserving structural integrity. The detailed process follows.}
}

%pipeline
% \ZY{\subsubsection{Pixel-wise Fine-Grained Constraint}}
%deviation projection
% \ZY{Despite the effectiveness of mesh-based retargeting, undesirable distortion still occurs. This is because mesh-level constraints cannot be properly aligned with pixel-level saliency. Accordingly, we present a new approach to realize the mapping from mesh-wise loss to pixel-level loss.}
% To this end, we perform the alignment in the output pixel space，
% As illustrated in Fig.~\ref{alignment}. 
\begin{figure*}[!t]
	\centering
	\includegraphics[width=\linewidth]{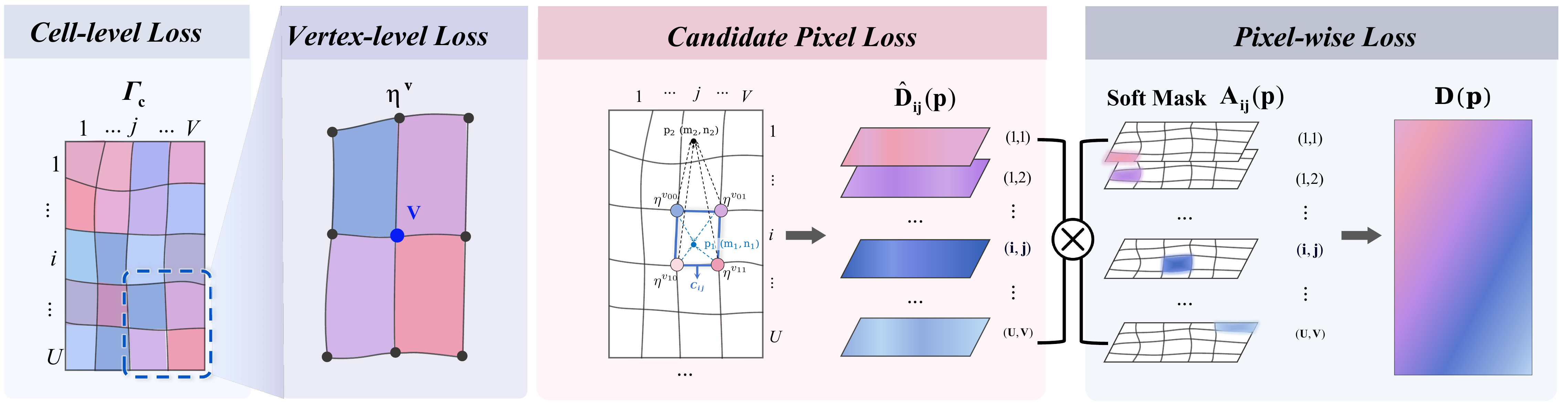}
	\caption{Flowchart of \ZY{building the pixel-wise constraints.}
   The cell-level mesh loss is first transformed into vertex-level values by averaging the losses over adjacent cells.
    These vertex values are subsequently interpolated within each deformed cell to derive candidate pixel-level constraints, and a soft mask is employed to select valid cell contributions.}
	\label{alignment}
\end{figure*}

% The alignment process consists of two branch: deviation projection branch and visual attention warping branch.
% Given the rigid mesh \(M^r\) and the deformed mesh \(M\), a geometric constraint first assigns each mesh cell \(\mathcal{C}_{ij}\) a scalar deviation value \(\delta_{ij}\). 

% For each type of geometric deviation, the corresponding cell-level values form a low-resolution cell-wise deviation field:
% \begin{equation}
% \Delta=\{\delta_{ij}\}_{i=1,j=1}^{U,V}. 
% \end{equation}

% \subsubsection{Deviation Projection}
\ZY{In this paper, instead of directly constraining the scaling and deformation of salient regions with pixel-level saliency, we convert cell-level loss of the \ZFL{vertex displacement} regression network to the pixel domain via interpolation over the mesh cell\ZZ{s}, and impose fine-grained constraints on salient regions in a content-aware manner.
As demonstrated in Fig.~\ref{alignment}, 
with the cell-level loss of the regression network defined in Sec.~\ref{network}, 
we first calculate the loss value on each vertex
according to its adjacent meshes, and then derive the per-pixel loss inside each cell by interpolating values from the corresponding vertices. The detailed implementation is elaborated as follows:
}

\ZY{For each vertex $v$ of the mesh, we search a set of mesh cells denoted as $\mathcal{N}(v)$ that share $v$ in the mesh topology, and the loss on $v$ is defined as:
}
% Let $\mathcal{N}(v)$ denote the set of mesh cells that share vertex $v$ in the mesh topology.
% The vertex-level deviation sample at $v$ is computed as：
\begin{equation}
\label{eq:vertexLoss}
    \eta^{v}
    =
    \frac{1}{|\mathcal{N}(v)|}
    \sum_{\mathcal{C} \in \mathcal{N}(v)}
    \Gamma_{\mathcal{C}},
\end{equation}
where $\Gamma_{\mathcal{C}}$ denotes the loss value of each neighboring mesh cell $\mathcal{C}$, 
\ZY{which is determined by the Saliency-guided Scale Preservation and Saliency-guided Structure Preservation defined in Sec.~\ref{loss}.}

% Given the deformed mesh $M$, each geometric constraint defines a cell-level deviation value for mesh cell $\mathcal{C}$. 
% The concrete formulation of these deviation values differs for the two specific constraint: \textit{Saliency-Aware Content-Preserving} and \textit{Global Structure-Preserving}.
% Details will be elaborated in section~\ref{loss}.

% To map cell-level geometric deviation values onto individual pixels of the output image, we first convert the cell-wise deviation into vertex-level deviation samples. 
% Vertex-level deviation values are averaged from neighboring cells deviation values.
% The aggregation is purely topological: it only uses the adjacency relation between cells and their shared vertices.
% Specifically, for each mesh vertex, we collect the deviation values of all cells that are topologically adjacent to this vertex and average them as its vertex-level deviation sample.
% Let $\mathcal{N}(v)$ denote the set of mesh cells that share vertex $v$ in the mesh topology.
% The vertex-level deviation sample at $v$ is computed as：
% \begin{equation}
%     \eta^{v}
%     =
%     \frac{1}{|\mathcal{N}(v)|}
%     \sum_{\mathcal{C}_k \in \mathcal{N}(v)}
%     \Gamma_{\mathcal{C}_k}，
% \end{equation}
% where $\Gamma_{\mathcal{C}_k}$ is the cell-level deviation value of each neighboring mesh cell $\mathcal{C}_k$ which is defined by two specific constraint: Saliency-Aware Content-
% Preserving and Global Structure-Preserving.

\ZFL{We map the vertex-wise losses to the pixel level using bilinear interpolation within each mesh cell.}
\ZY{Specifically, let $\hat{\mathcal{D}}_{ij}$ denote the candidate loss interpolated from grid cell $\mathcal{C}_{ij}$,} where $i=1,\ldots,U$ and $j=1,\ldots,V$.
%% TODO
For each pixel $p$ in the target output domain, we compute its normalized local coordinates $(m,n)$ with respect to $\mathcal{C}_{ij}$.
\ZY{As illustrated in Fig.~\ref{alignment}, $\hat{\mathcal{D}}_{ij}(p)$ denotes the set of all pixel loss candidates interpolated by each individual mesh cell.}
% each mesh cell $\{\mathcal{C}_{i,j}| i \in [1,U], j \in[1,V]\}$.
% As shown in Fig.~\ref{softmask}, let $\eta^{v_{00}}$, $\eta^{v_{01}}$, $\eta^{v_{11}}$, and $\eta^{v_{10}}$ denote the vertex-level losses associated with the four vertices of each mesh cell \ZY{$\{\mathcal{C}_{i,j}\}$}.
Accordingly, the candidate loss contributed by each mesh cell is then computed as:
\begin{equation}
\begin{aligned}
    \hat{\mathcal{D}}_{ij}(p)
    =
    &(1-m)(1-n)\cdot\eta^{v_{00}}
    + m(1-n)\cdot\eta^{v_{01}} \\
    &+ mn\cdot \eta^{v_{11}}
    + (1-m)n\cdot\eta^{v_{10}}.
\end{aligned}
\label{bilinear}
\end{equation}
where $\eta^{v_{00}}$, $\eta^{v_{01}}$, $\eta^{v_{11}}$, and $\eta^{v_{10}}$ denote the vertex-level loss at the four vertices of each mesh cell. 

%why softmask
% Although eq.~\ref{bilinear} estimates candidate deviation from each deformed mesh cell, it does not determine the valid region (e.g. the pixel $p$ may outside of the current deformed mesh cell $c_{ij}$).
% To restrict each candidate deviation to the actual coverage of its corresponding deformed mesh cell, we introduce cell-wise soft masks.

% Although Eq.~\ref{bilinear} estimates a candidate loss from each deformed mesh cell, this candidate should only contribute to pixels covered by that cell in deformed mesh.

\begin{figure}[!t]
	\centering
	\includegraphics[width=\linewidth]{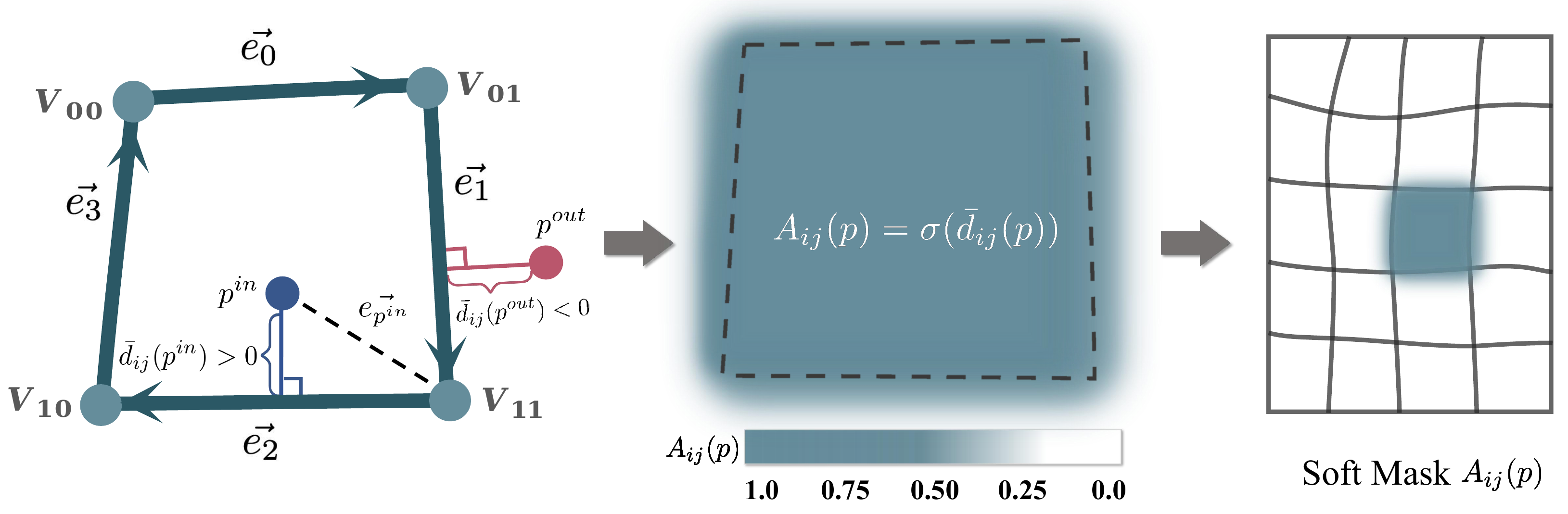}
	\caption{Illustration of the soft mask $\mathcal{A}_{ij}(p)$.
    For a grid cell $\mathcal{C}_{ij}$, the normalized minimum signed distance is defined as $\bar d_{ij}(p)=\min_{t\in\{0,1,2,3\}}\frac{\vec e_t \times \vec e_p}{\|\vec e_t\|_2}$, where $\vec e_t$ denotes an oriented cell edge, and $\vec e_{p}$ denotes the vector connecting the starting vertex of $\vec e_t$ to pixel $p$.
    Pixels inside and outside the cell are assigned positive and negative signed distances, respectively. 
    The $\operatorname{Sigmoid}$ function maps $\bar d_{ij}(p)$ to the corresponding soft mask weight $A_{ij}(p)$.}
	\label{softmask}
\end{figure}

\ZY{Since $\hat{\mathcal{D}_{ij}}$ provides the loss values interpolated by an individual cell $\mathcal{C}_{ij}$, only pixel losses within \(\mathcal{C}_{ij}\) are valid. 
We therefore further design a mask function to extract the effective losses.
% To improve the efficiency, Eq.~\ref{bilinear} parallelly calculates the interpolated values across all mesh cells for each pixel. Each pixel value is thus represented as a column vector whose dimension corresponds to the number of mesh cells.
% Accordingly, we need to design an effective mask function to select the correct interpolation result from all candidates for each pixel.
% The main idea of this function is to evaluate the spatial relation between each pixel and every mesh cell, outputting `1' if the pixel is inside the cell and `0' otherwise. 
The main idea of this function is to evaluate the spatial relation between each pixel and every mesh cell,
assigning a soft weight approaching `1' to pixels inside the cell and approaching `0' to pixels outside the cell. 
% Concretely, signed distances from the pixel to the four boundaries of each cell are calculated and normalized with the Sigmoid function.
Concretely, signed distances from the pixel to the four boundaries of each cell are calculated, and the normalized minimum signed distance is mapped to a soft weight by the Sigmoid function. 
We define the function as follows:
}
% We view this validity as binary masking determined by the relative position between an output pixel $p \in Y$ and the corresponding deformed mesh cell $\mathcal{C}$. 
% %define&formula
% To make the mapping continuous and differentiable, we adopt a cell-wise soft mask:
\begin{equation}
    \mathcal{A}_{ij}(p)
    =\sigma
    \left(
    \bar{d}_{ij}(p)
    \right),
\end{equation}
where $\sigma$ refers to the \text{Sigmoid} function, and $\bar{d}_{ij}(p)$ denotes the normalized minimum signed distance from pixel $p$ to the edges of the grid cell $\mathcal{C}_{ij}$.
The positive and negative values indicate pixels inside and outside the cell, respectively.
As illustrated in Fig.~\ref{softmask}, 
% the Sigmoid function converts the signed distance into continuous weight.
\ZY{the signed distance for each pixel has a dimension equal to the number of mesh cells. By simply extracting each dimension, we can conveniently obtain the cell-wise soft mask.}

\ZY{Finally, we calculate the pixel-wise loss using the following function:}
% Given the candidate pixel losses $\hat{D}(p)$ and the soft mask $A(p)$, the final pixel-wise loss is defined as:
\begin{equation}
\label{eq:pixlewiseloss}
    \mathcal{D}(p)
    =
    \frac{\sum_{i=1}^{U}
    \sum_{j=1}^{V}
    \mathcal{A}_{ij}(p)
    \hat{\mathcal{D}}_{ij}(p)}{\sum_{i=1}^{U}
    \sum_{j=1}^{V}
    \mathcal{A}_{ij}(p)},
\end{equation}
\ZY{where $i, j$ denote the row and column indices of mesh cells, respectively.}

% The resultant dense map $D$ assigns each output pixel $p\in\Omega'$ a geometric deviation measure $D(p)$,  which reflects the local mesh-induced geometric deviation inherited from its surrounding cells. 
% Compared with directly assigning saliency weights to coarse mesh cells, this projection provides a dense geometric description that is spatially compatible with the retargeted image.

% \subsubsection{Visual Attention Warping}
% Meanwhile, the saliency map $S$ is originally defined in the input image space. 
% We therefore warp $S$ with the same deformed mesh $M$:
% \begin{equation}
% S_{warp}=\mathcal{W}(S;M),
% \end{equation}
% where $\mathcal{W}(\cdot;M)$ denotes the mesh-guided warping. 
% After this operation, $S_{warp}$ and $D^k(p)$ are both pixel level. 
%  This common pixel-domain representation enables pixel-wise alignment between dense visual attention and geometric deviation.

\subsection{Objective functions}
\label{loss}
% We optimize our network parameters using a comprehensive objective function consisting of three terms.
% The total objective is formulated as:
\ZY{
We optimize network parameters with a comprehensive three-term objective function formulated as:
}
\begin{equation}
    \mathcal{L}_{total}=
    \lambda_c\mathcal{L}_{c}
    +
    \lambda_t\mathcal{L}_{t}
    +
    \lambda_g\mathcal{L}_{g},
\end{equation}
where $\mathcal{L}_{c}$ denotes the saliency-guided scale preservation, $\mathcal{L}_{t}$ denotes the saliency-guided structure preservation, and $\mathcal{L}_{g}$ denotes the mesh geometry preservation. $\lambda_c$, $\lambda_t$, and $\lambda_g$ denote the corresponding weights.

\subsubsection{Saliency-guided Scale Preservation}
%motivation
\ZY{Given the saliency map $\mathcal{S}$ of the input image, our saliency-aware content-preserving loss enforces proper scaling and suppresses distortion for salient regions during retargeting.}

% Given the predicted mesh $\mathcal{M}$, we encourage local scale consistency in salient regions while allowing less-attended regions to undergo more flexible deformation.
%design

To achieve this goal, 
\ZY{we first define a cell-level loss to enforce content-aware scaling in salient regions, then we further map the cell-wise loss to pixel-level for fine-grained constraint, according to the solution defined in Sec.~\ref{finegrain}.}
% we define a local scale deviation for each mesh cell and penalize it according to the dense visual attention distribution.
For each mesh cell, we measure its horizontal and vertical scaling factors \ZY{as}:
\begin{equation}
s^{x}=\frac{V}{W}w,
\qquad
s^{y}=\frac{U}{H}h,
\end{equation}
where $w$ and $h$ denote the horizontal and vertical \ZY{dimensions of the deformed mesh cell, and $W$ and $H$ denote the dimensions of the input image.}
The loss value of each cell is then defined as:
\begin{equation}
    \Gamma_{\mathcal{C}}
    =
    (s^x-1)^2+(s^y-1)^2.
    \end{equation}

\ZY{With the cell-level constraint $\Gamma_{\mathcal{C}}$,
we further define the pixel-wise loss $\mathcal{D}(p)$ according to the equations defined from Eq.~\ref{eq:vertexLoss} to Eq.~\ref{eq:pixlewiseloss}}, and the saliency-aware content-preserving loss is formulated as:
\begin{equation}
\mathcal{L}_c
=
\sum_{p\in Y}
\mathcal{W}(\mathcal{S}, \mathcal{M})
\mathcal{D}(p),
\end{equation}
where $\mathcal{W}(\mathcal{S}, \mathcal{M})$ denotes the saliency map obtained by warping the input saliency map $\mathcal{S}$ according to the deformed mesh $\mathcal{M}$.

\subsubsection{Saliency-guided Structure Preservation}
Although the saliency-guided scale preservation term constrains local scale variations in salient regions, the mesh cells may still undergo irregular deformation to accommodate aspect-ratio changes.
Such unconstrained deformation can introduce shear, rotation, or bending artifacts in the retargeted image. 
We \ZZ{therefore} introduce a saliency-guided structure preservation term to suppress irregular mesh deformation.
% To prevent irregular geometric deformation in the retargeted image, the deformed mesh should not be distorted in an uncontrolled manner. Therefore, we design an intra-cell orientation constraint and an inter-cell motion coherence constraint to keep the structural regularity of the deformed mesh.
For each mesh cell $\mathcal{C}$ in the deformed mesh $M$, we denote its four edges as $\mathbf{e}$, which contains two horizontal edges $\mathbf{e}^h$ and two vertical edges $\mathbf{e}^v$.
% For each edge $\mathbf{e}\in\mathcal{E}$, we assign an expected unit direction $\mathbf{t}_{\mathbf{e}}$, where $\mathbf{t}^h_{\mathbf{e}}=(1,0)^{\top}$ for horizontal edges and $\mathbf{t}^v_{\mathbf{e}}=(0,1)^{\top}$ for vertical edges.
The orientation deviation of edge $\mathbf{e}$ is defined as:
\begin{equation}
\psi(\mathbf{e})
=
1-
\frac{
(\mathbf{e}^{\top}\mathbf{t}_{\mathbf{e}})^2
}{
\|\mathbf{e}\|_2^2
},
\end{equation}
where $\mathbf{t}_{\mathbf{e}}$ denotes the expected unit orientation of edge $\mathbf{e}$, with $\mathbf{t}_{\mathbf{e}}=(1,0)^{\top}$ for horizontal edges and $\mathbf{t}_{\mathbf{e}}=(0,1)^{\top}$ for vertical edges.
Then, the \ZY{cell-wise} structural loss $\Gamma_{\mathcal{C}}$ is computed by averaging the orientation deviations of its four edges:
\begin{equation}
\Gamma_{\mathcal{C}}
=
\frac{1}{4}
\sum_{\mathbf{e}\in\mathcal{E}}
 \psi(\mathbf{e}).
\end{equation}

\ZY{Similar to the scale preservation loss defined above, we further map the mesh-wise loss to pixel-level using the equations defined from Eqs.~\ref{eq:vertexLoss} to~\ref{eq:pixlewiseloss}
and the pixel-wise loss $\mathcal{D}(p)$ is adopted to impose global structural constraint, the definition is as follows:
}
% Following Sec.~\ref{finegrain}, $\Gamma^{g}_{\mathcal{C}}$ is mapped onto the pixel domain, yielding the pixel-wise structural loss $\mathcal{D}^g(p)$.
% We then modulate $\mathcal{D}^g(p)$ using the same $\mathcal{S}_{warp}$, imposing stronger structural regularization on salient regions and weaker regularization on less-salient regions.
% The intra-cell orientation constraint is formulated as:

\begin{equation}
\mathcal{L}_{t}
=
\sum_{p\in Y}
\left[
\mathcal{W}(\mathcal{S}, \mathcal{M})
+
\beta
\left(
1-\mathcal{W}(\mathcal{S}, \mathcal{M})
\right)
\right]
\mathcal{D}(p),
\end{equation}
where $\beta=0.5$ regulates the degree of structural preservation imposed on less salient regions.
\begin{figure}[!t]
	\centering
	\includegraphics[width=0.94\linewidth]{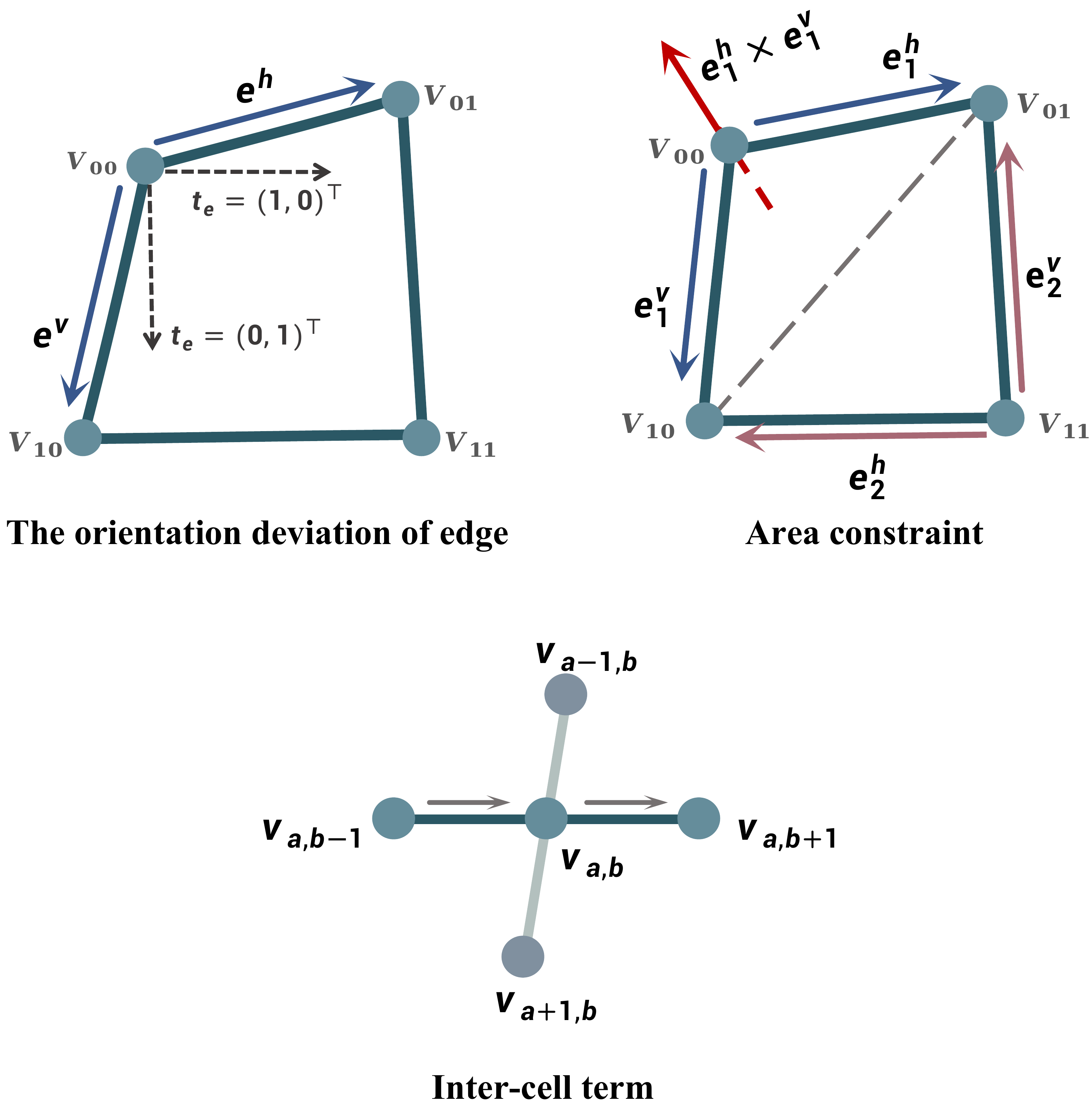}
	\caption{Mesh geometry preservation constraint, which consists of intra-cell and inter-cell terms.}
	\label{constraint}
\end{figure}

% {\textit{Inter-cell motion coherence constraint.}}
% To encourage neighboring vertices to move coherently, we impose a smoothness regularization on motion $\Delta M$.
% We denote the motion of vertex $v_{a,b}$ as $\Delta M_{a,b}$, where $a\in[0,U]$ and $b\in[0,V]$.
% The inter-cell motion coherence constraint is formulated as:
% \begin{equation}
% \begin{aligned}
%     \mathcal{L}_{inter}
%     =
%     \frac{1}{(U+1)\cdot V}
%     \sum_{a=0}^{U}
%     \sum_{b=0}^{V-1}
%     \left\|
%     \Delta M_{a,b+1}
%     -
%     \Delta M_{a,b}
%     \right\|_2^2 \\
%     +
%     \frac{1}{U\cdot(V+1)}
%     \sum_{a=0}^{U-1}
%     \sum_{b=0}^{V}
%     \left\|
%     \Delta M_{a+1,b}
%     -
%     \Delta M_{a,b}
%     \right\|_2^2.
% \end{aligned}
% \end{equation}

\subsubsection{Mesh Geometry Preservation}
% {\textbf{Mesh Geometry Preservation.}}
\ZY{To maintain the geometric validity and smoothness of the mesh cells during deformation, we design both \textit{intra-cell} and \textit{inter-cell} constraints. The former suppresses flipping and folding via local edge and area metrics, while the latter enforces a smoothness regularization to encourage coherent motion among neighboring cells.}
% \ZY{To keep the mesh cells geometrically valid and smooth during deformation, we suppress flipping and folding via local edge and area constraints,} and impose a smoothness regularization to encourage neighboring vertices to move coherently.
% To preserve local shape, we aim to prevent folding and flipping of individual mesh cells.
% We propose to constrain the local mesh shape from two perspectives: edge and area.

\textbf{Intra-cell term.}
%edge
\ZY{For each mesh cell, we first enforce the direction and magnitude of each edge to avoid flipping and folding.
Similar to~\cite{DBLP:conf/cvpr/NieLLL022},
we also encourage horizontal edges to project toward the right and vertical edges to project downward, while imposing constraints on their lengths.
For each horizontal edge $\mathbf e\in\mathcal{E}^{h}$ and vertical edge $\mathbf e\in\mathcal{E}^{v}$, following the formulation in~\cite{DBLP:conf/cvpr/NieLLL022}, we construct penalty terms $\mathcal{P}_h$ and $\mathcal{P}_v$ to mitigate cell flipping and folding,} and the edge constraint is defined as:
\begin{equation}
    \mathcal{L}_{edge}
    =
    \sum_{i=1}^{U}
    \sum_{j=1}^{V}
    \left(
    \sum_{\mathbf e\in\mathcal{E}^{h}_{ij}} \mathcal{P}_h(\mathbf e)
    +
    \sum_{\mathbf e\in\mathcal{E}^{v}_{ij}} \mathcal{P}_v(\mathbf e)
    \right).
\end{equation}

% Following the edge notation defined above, we encourage the projection of each horizontal edge $\mathbf e\in\mathcal{E}^{h}$ along the expected direction $\mathbf{t}^h_{\mathbf{e}}$ to be at least a threshold $\alpha \frac{W}{V}$.
% We use a penalty $P_h$ to describe this constraint as follows:
% \begin{equation}
%     P_h(\mathbf e)
%     =
%     \begin{cases}
%     \left(
%     \alpha \dfrac{W}{V} - \mathbf e^{\top}\mathbf t^h_{\mathbf e}
%     \right)^2,
%     &
%     \mathbf e^{\top}\mathbf t^h_{\mathbf e} < \alpha \dfrac{W}{V},
%     \\
%     0,
%     &
%     \mathbf e^{\top}\mathbf t^h_{\mathbf e} \geq \alpha \dfrac{W}{V},
%     \end{cases}
% \end{equation}
% where $\alpha=0.2$ is introduced to prevent excessive compression and flipping.
% As for vertical edge $\mathbf e\in\mathcal{E}^{v}$, we impose a similar penalty $P_v$ as follows:
% \begin{equation}
%     P_v(\mathbf e)
%     =
%     \begin{cases}
%     \left(
%     \alpha \dfrac{H}{U} - \mathbf e^{\top}\mathbf t^v_{\mathbf e}
%     \right)^2,
%     &
%     \mathbf e^{\top}\mathbf t^v_{\mathbf e} < \alpha \dfrac{H}{U},
%     \\
%     0,
%     &
%     \mathbf e^{\top}\mathbf t^v_{\mathbf e} \geq \alpha \dfrac{H}{U}.
%     \end{cases}
% \end{equation}

% \textbf{Inter-cell term.}
% This constraint prevents neighboring vertices from becoming overly close along the horizontal and vertical directions. 

To further prevent the \ZY{geometric} degeneration of mesh cells, such as flipping, we \ZY{impose} an area constraint on each mesh cell.
As illustrated in Fig.~\ref{constraint}, each mesh cell $\mathcal{C}$ is decomposed into two triangles.
For \ZY{each} mesh cell, we define the area penalty as:
\begin{equation}
\ZY{\mathcal{P}_r(\mathcal{C})} = 
\begin{cases} 
\ZY{\sum_{r=1}^{2}}(A_{min} - \mathbf{e}_r^h \times \mathbf{e}_r^v)^2 & \mathbf{e}_r^h \times \mathbf{e}_r^v < A_{min} \\
0, & \mathbf{e}_r^h \times \mathbf{e}_r^v \ge A_{min} 
\end{cases},
\end{equation}
where $r \in \{1, 2\}$, $\mathbf{e}_r^h$ and $\mathbf{e}_r^v$ denote the directed horizontal and vertical mesh edges of a triangle in the mesh, respectively.
$\mathbf{e}_r^h \times \mathbf{e}_r^v$ \ZY{denotes the signed area formed by adjacent edges. Specifically, a signed area near zero means the corresponding triangle degenerates into a line segment, whereas a negative value indicates flipped triangle geometry.}
 % Non-positive values correspond to cell flipping}, and
$A_{min}=0.2 \cdot\frac{W}{V}\cdot\frac{H}{U}$ \ZY{defines} the minimum area threshold.
% When the signed area approaches zero, the corresponding triangle degenerates into a line, when it becomes negative, the triangle is flipped.
The area constraint is then formulated as:
\begin{equation}
\mathcal{L}_{area} = \sum_{i=1}^{U} \sum_{j=1}^{V} \ZY{\mathcal{P}_r(\mathcal{C}_{ij})}.
\end{equation}

\textbf{Inter-cell term.}
\ZY{
The intra-cell term defined above ensures geometric correctness inside each mesh cell, yet it cannot produce smooth transitions between cells for natural deformation results.
% To enforce coherent displacement of adjacent vertices and globally smooth deformation for natural retargeting,
To maintain smooth \ZFL{vertex displacement} and uniform displacements among adjacent vertices, we constrain the discrete second derivatives of the mesh vertices, defined as follows:
}
% we impose a smoothness regularization on motion $\Delta M$.
% We denote the motion of \ZY{adjacent} vertices as $\Delta M_{a,b} = v_b - v_a$, where $a\in[0,U]$ and $b\in[0,V]$.
% The smoothness regularization is formulated as:
\begin{equation}
\begin{aligned}    
\mathcal{L}_{smooth}    
=    
\frac{\sum\limits_{a=0}^{U}    
\sum\limits_{b=0}^{V-1}    
\left\|    
v_{a-1,b} - 2\cdot v_{a,b} + v_{a+1,b}      
\right\|_2^2}{(U+1)\cdot V}    
 \\    
+    
\frac{\sum\limits_{a=0}^{U-1}    
\sum\limits_{b=0}^{V}    
\left\|    
v_{a,b-1} - 2\cdot v_{a,b} + v_{a,b+1}    
\right\|_2^2}{U\cdot(V+1)}.
\end{aligned}
\end{equation}

In sum, the total mesh geometry loss is formulated as:
\begin{equation}
    \mathcal{L}_{g} = \mathcal{L}_{edge} + \lambda_a \mathcal{L}_{area} + \mathcal{L}_{smooth},
\end{equation}
where $\lambda_a=0.8$ is set to balance the contributions of these terms.

% {\textbf{smoothness???}}
% Although the above constraints preserve salient structures and prevent unintended mesh deformation, they do not guarantee global geometric smoothness.
% Therefore, we impose a smoothness regularization on motion $\Delta M$ to encourage neighboring vertices to move coherently.

% Specifically, we denote the motion of vertex $v_{a,b}$ as $\Delta M_{a,b}$, where $a\in[0,U]$ and $b\in[0,V]$.

\section{Experiments}
\subsection{Implementation Details}
Our framework is implemented in PyTorch on a single NVIDIA A40 GPU.
During training, all input images are resized to $224 \times 224$ and normalized to $[-1,1]$.
The corresponding saliency maps are \ZY{obtained by the pretrained model from}~\cite{gao2024mdsam}, and normalized to $[0,1]$.
\ZY{Following the practice of prior work~\cite{cho2017weakly}, we randomly sample input scaling ratios from 1/4 to 1/2 for \ZZ{each} batch, which corresponds to scaling the original 224-pixel side length to a size between 224/4 and 224/2.
}
% input aspect ratios are randomly generated for each batch within the range of $224/4$ to $224/2$.
\ZY{During the training stage, we use} the Adam optimizer with an initial learning rate of $3\times 10^{-5}$, \ZY{and }the learning rate is decayed by a factor of 0.97.
\ZY{We train the network over 50 epochs with a batch size of 32 and ReLU as the activation function. }
% The network is trained for 50 epochs.
% The batch size is set to 32, and \ZY{ReLU} is used as the activation function.
To stabilize the optimization process under large mesh deformation, gradient clipping with a maximum norm of 3 is adopted during backpropagation.
\ZY{For the loss function, we set}
$\lambda_c$, $\lambda_t$, and $\lambda_g$ to 300, 50, and 1, respectively, and set $U \times V$ to $11 \times 11$.

\ZZ{Similar to~\cite{liao2026object}, we train the model only for size reduction; enlarged retargeting results are produced by applying inverse mesh deformation.} Specifically, for an enlargement target, we first estimate the deformation under the corresponding reduction setting and then reverse the mesh-guided warping direction to obtain the enlarged output.
% Similar to~\cite{liao2026object}, we train our model only in reducing size, it can also produce enlarged retargeting results by applying inverse mesh deformation. 
% Specifically, for an enlargement target, we first estimate the deformation under its corresponding reduction setting and then reverse the mesh-guided warping direction to obtain the enlarged output. 
% The detailed inverse deformation procedure follows ObjectIR~\cite{liao2026object}.

\ZY{We evaluate our model across two standard datasets for image retargeting tasks.
Most of our training and testing are performed
}on the Object-IR dataset~\cite{liao2026object}.
The dataset is \ZY{derived} from a filtered COCO dataset~\cite{lin2014microsoft}, which contains 18023 images in total, including 17043 images for training and 980 images for testing.
\ZY{We further }evaluate our method on the public image retargeting dataset RetargetMe~\cite{rubinstein2010comparative}, which contains 80 images spanning a wide range of scenes.
Many representative retargeting methods have reported retargeted results at different aspect ratios on RetargetMe, which makes direct comparison under this benchmark straightforward.
For methods with available source codes, we generate their results and conduct comprehensive comparisons using both quantitative and visual quality assessment.
\ZY{In the experiment, we compare our approach with several representative and state-of-the-art methods, including CR (Cropping),
SCL (Simple Scaling),
SC~\cite{avidan2007seam}, 
SNS~\cite{dong2009optimized}, 
WSSDCNN~\cite{cho2017weakly}, 
GDPM~\cite{elnekave2022generating},
Object-IR~\cite{liao2026object}.
}
%%%%%%%%%%%%%%%%%%%%%%%%%%%%%%%%%%%%%%

\begin{table*}[!t]
    \centering
    \caption{
    Quantitative comparisons with representative retargeting approaches on two datasets using the object-based distortion error metric from Object-IR~\cite{liao2026object}, including CR (Cropping), SCL (Simple Scaling), SC~\cite{avidan2007seam}, SNS~\cite{dong2009optimized}, WSSDCNN~\cite{cho2017weakly},
    GDPM~\cite{elnekave2022generating}, Object-IR~\cite{liao2026object}. Each cell is formatted as ``value on ObjectIR testing dataset~\cite{liao2026object} / value on RetargetMe dataset~\cite{rubinstein2010comparative}''. 
    \colorbox{pink}{Red} and \colorbox{cyan!35}{Cyan} colorboxes indicate the best performance on the ObjectIR~\cite{liao2026object} and RetargetMe~\cite{rubinstein2010comparative} datasets, respectively. Underlined values denote the second-best results, and lower values indicate superior retargeting quality.}
    \begin{tabular}{lccccc}
        \toprule
        {Methods} 
        & {0.5$\times$} 
        & {0.75$\times$} 
        & {1.25$\times$} 
        & {1.5$\times$} 
        & {1.75$\times$} \\
        \midrule
        CR 
        & 0.4652/0.7761 
        & 0.3251/0.5347 
        & --/-- 
        & --/-- 
        & --/-- \\
        SCL 
        & 0.6357/0.7721 
        & 0.3074/0.3507 
        & 0.2873/0.2531 
        & 0.5323/0.5392 
        & 0.8096/0.7096 \\
        SC~\cite{avidan2007seam} 
        & 0.4054/0.5615 
        & 0.2984/\underline{0.2944} 
        & 0.2597/\underline{0.2182} 
        & \underline{0.3352}/0.4945 
        & 0.6042/0.6991 \\
        SNS~\cite{dong2009optimized} 
        & \underline{0.3280}/0.6689 
        & 0.2941/0.4677 
        & 0.2550/0.3993 
        & 0.4901/0.4971 
        & 0.4824/\underline{0.4853} \\
        WSSDCNN~\cite{cho2017weakly} 
        & 0.7934/0.7304 
        & 0.5268/0.4249 
        & --/-- 
        & --/-- 
        & --/-- \\
        GDPM~\cite{elnekave2022generating} 
        & 0.8658/0.8158 
        & 0.6828/0.7113 
        & 0.4878/0.4393 
        & 0.4974/0.6837 
        & \underline{0.4572}/0.6243 \\
        ObjectIR~\cite{liao2026object} 
        & 0.4074/\underline{0.5434} 
        & \colorbox{pink}{0.2114}/0.3080 
        & \underline{0.2440}/0.2514 
        & 0.4666/\underline{0.4831} 
        & 0.6794/0.6945 \\
        Ours 
        & \colorbox{pink}{0.2917}/\colorbox{cyan!35}{0.4420} 
        & \underline{0.2271}/\colorbox{cyan!35}{0.2845} 
        & \colorbox{pink}{0.1758}/\colorbox{cyan!35}{0.2004} 
        & \colorbox{pink}{0.3178}/\colorbox{cyan!35}{0.4453} 
        & \colorbox{pink}{0.3400}/\colorbox{cyan!35}{0.4273} \\
        \bottomrule
    \end{tabular}
    
    \label{tab:overall_quantitative}
\end{table*}

\begin{table}[t]
\centering
\caption{Additional perceptual quality comparison with PruneRepaint~\cite{shen2024prune} on Retargetme~\cite{rubinstein2010comparative} dataset. 
DISTS and MUSIQ are used to measure perceptual distortion and no-reference image quality, respectively.}
\setlength{\tabcolsep}{18pt}
\begin{tabular}{lcc}
\toprule
{Methods} & {DISTS} $\downarrow$ & {MUSIQ} $\uparrow$ \\
\midrule
Prune~\cite{shen2024prune} & $0.2016$ & $65.56$ \\
Ours & $\mathbf{0.1429}$ & $\mathbf{70.79}$ \\
\bottomrule
\end{tabular}

\label{tab:prunerepaint_quality}
\end{table}
%%%%%%%%%%%%%%%%%%%%%%%%%%%%%%%%%%%%%%

\subsection{Quantitative Comparison}
\ZY{We first perform quantitative comparison using the }
object distortion metric proposed in~\cite{liao2026object}\ZY{, where lower values indicate better retargeting performance}.
\ZY{We evaluate the retargeting performance at five distinct scales:}
$0.5$, $0.75$, $1.2$, $1.5$ and $1.75$.
\ZY{To achieve comprehensive quantitative assessment, we conduct evaluations on both the Object-IR~\cite{liao2026object} and RetargetMe~\cite{rubinstein2010comparative} datasets, as reported in Table~\ref{tab:overall_quantitative}.}
\ZZ{The symbol} \ZY{``--''} indicates that the corresponding method cannot generate retargeted results under the given resizing ratio. 
The best and second-best results are marked in bold and underlined, respectively.
\ZY{For the CR and WSSDCNN~\cite{cho2017weakly} baselines, only image downscaling is supported, so we do not include upscaling measurements for these two methods.}
% Compared with previous methods, our distortion scores are more convincing across different scales.
\ZY{As shown in Table~\ref{tab:overall_quantitative}, our distortion scores outperform prior methods across diverse scaling settings on different datasets. 
Other content-aware retargeting methods, such as SC~\cite{avidan2007seam}, SNS~\cite{dong2009optimized}, and ObjectIR~\cite{liao2026object}, can mitigate object-level distortion to a certain degree, yet their performance fluctuates drastically under varying resizing ratios.
Specifically, 
ObjectIR~\cite{liao2026object} only attains a marginal performance gain at the scaling factor of 0.75.
In contrast, our method maintains superior performance throughout all scales, fully validating its excellent adaptability and robustness.}

\begin{figure*}[!t]
	\centering
	\includegraphics[width=\linewidth]{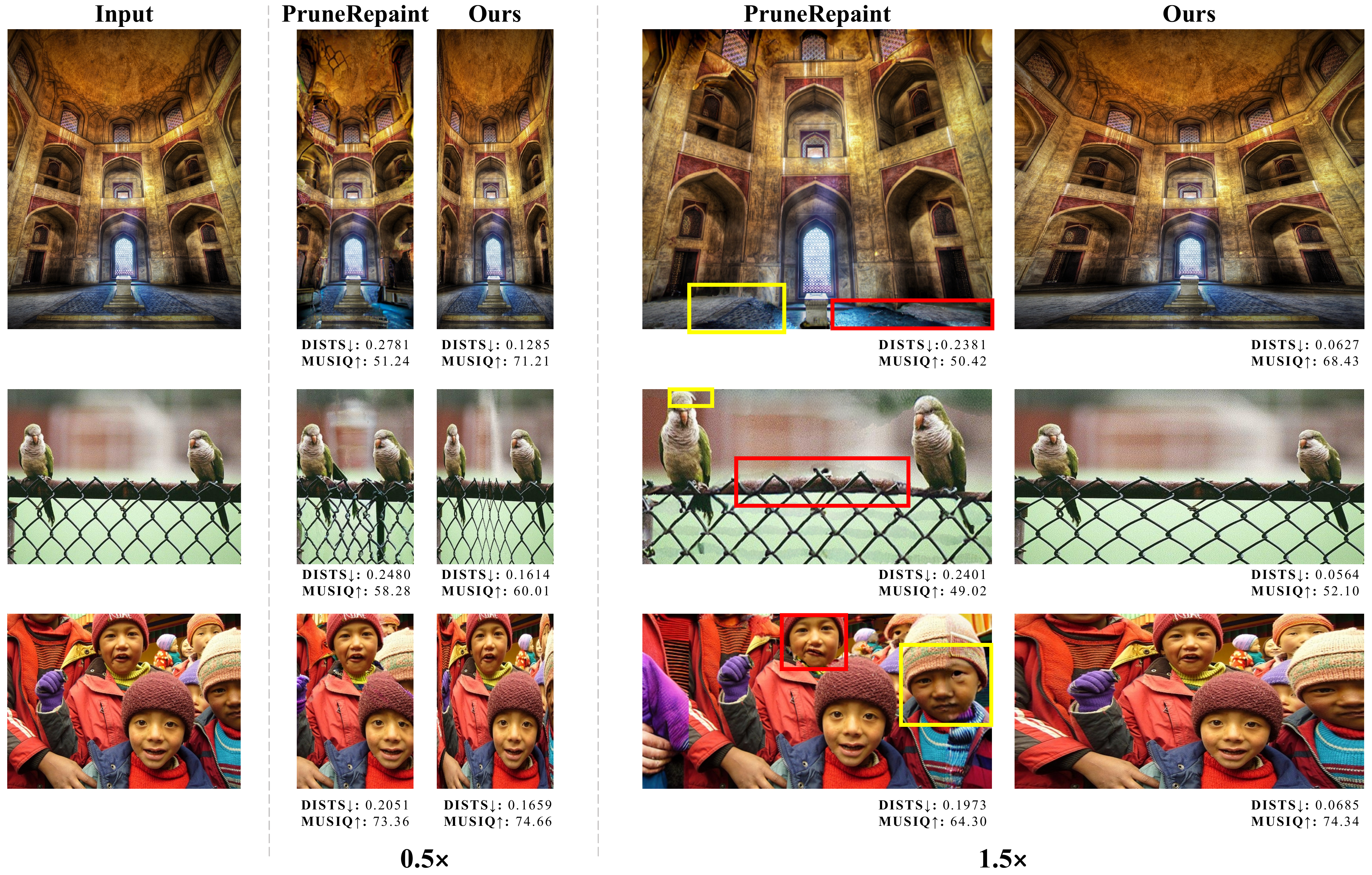}
	\caption{Visual comparison with PruneRepaint~\cite{shen2024prune}. DISTS and MUSIQ are reported below each retargeted result.
    Lower DISTS and higher MUSIQ \ZY{values} indicate better perceptual consistency and overall image quality, respectively.}
\label{visualCompare_prunerepaint}
\end{figure*}

\ZY{Table~\ref{tab:prunerepaint_quality} provides extensive comparisons with generative learning-based methods via two comprehensive perceptual metrics: DISTS~\cite{ding2020image} and MUSIQ~\cite{ke2021musiq}.
Specifically,} DISTS indicates perceptual distortion between the retargeted image and the reference image, and  MUSIQ refers to a no-reference image quality assessment metric.
As shown in Table~\ref{tab:prunerepaint_quality},  our method achieves a lower DISTS score and a higher MUSIQ score\ZY{, which}
indicates that our method produces less perceptual distortion and better overall visual quality. 
The visual examples in Fig.~\ref{visualCompare_prunerepaint} further demonstrate that our method better preserves original image content and avoids artifacts.

\begin{figure*}[!t]
	\centering
	\includegraphics[width=\linewidth]{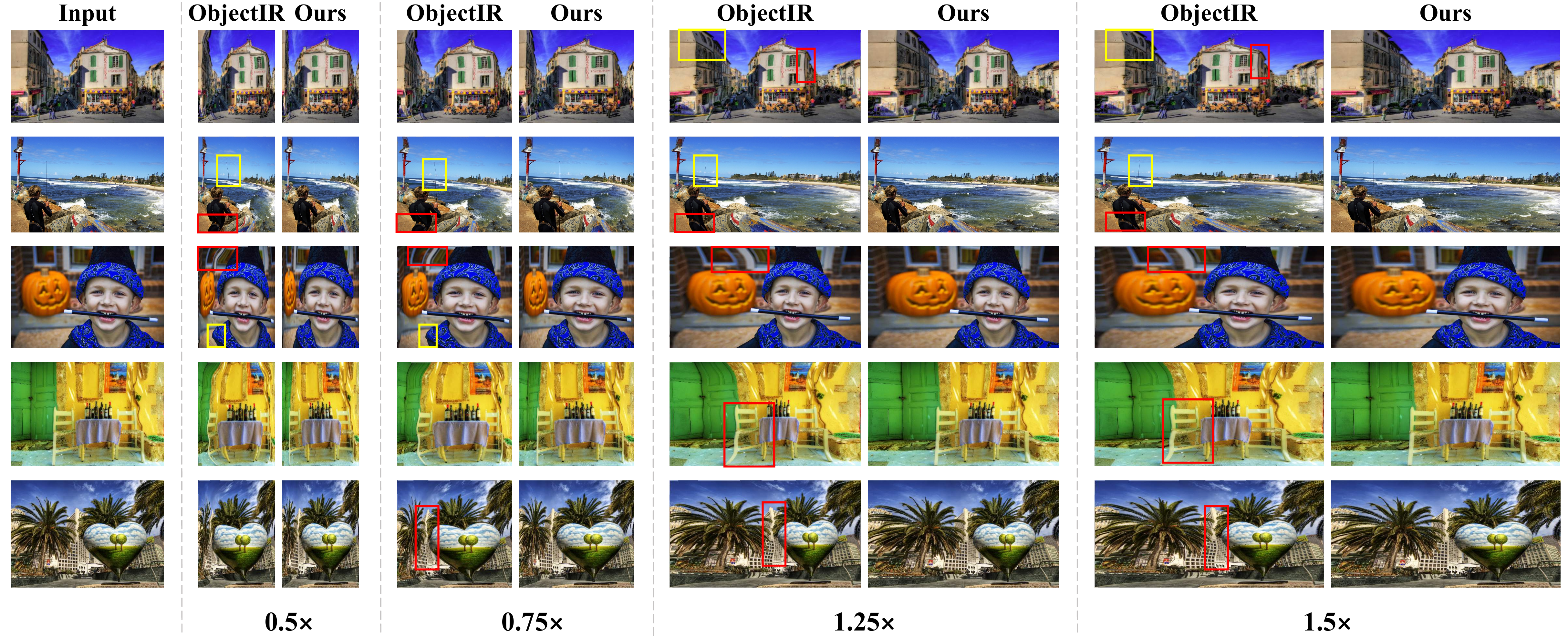}
    \caption{Visual comparison with the SOTA retargeting method~\cite{liao2026object} across different scales\ZY{, where colored boxes mark the distorted artifacts of~\cite{liao2026object}.}}
\label{differentScale}
\end{figure*}

\begin{figure*}[!t]
	\centering
	\includegraphics[width=\linewidth]{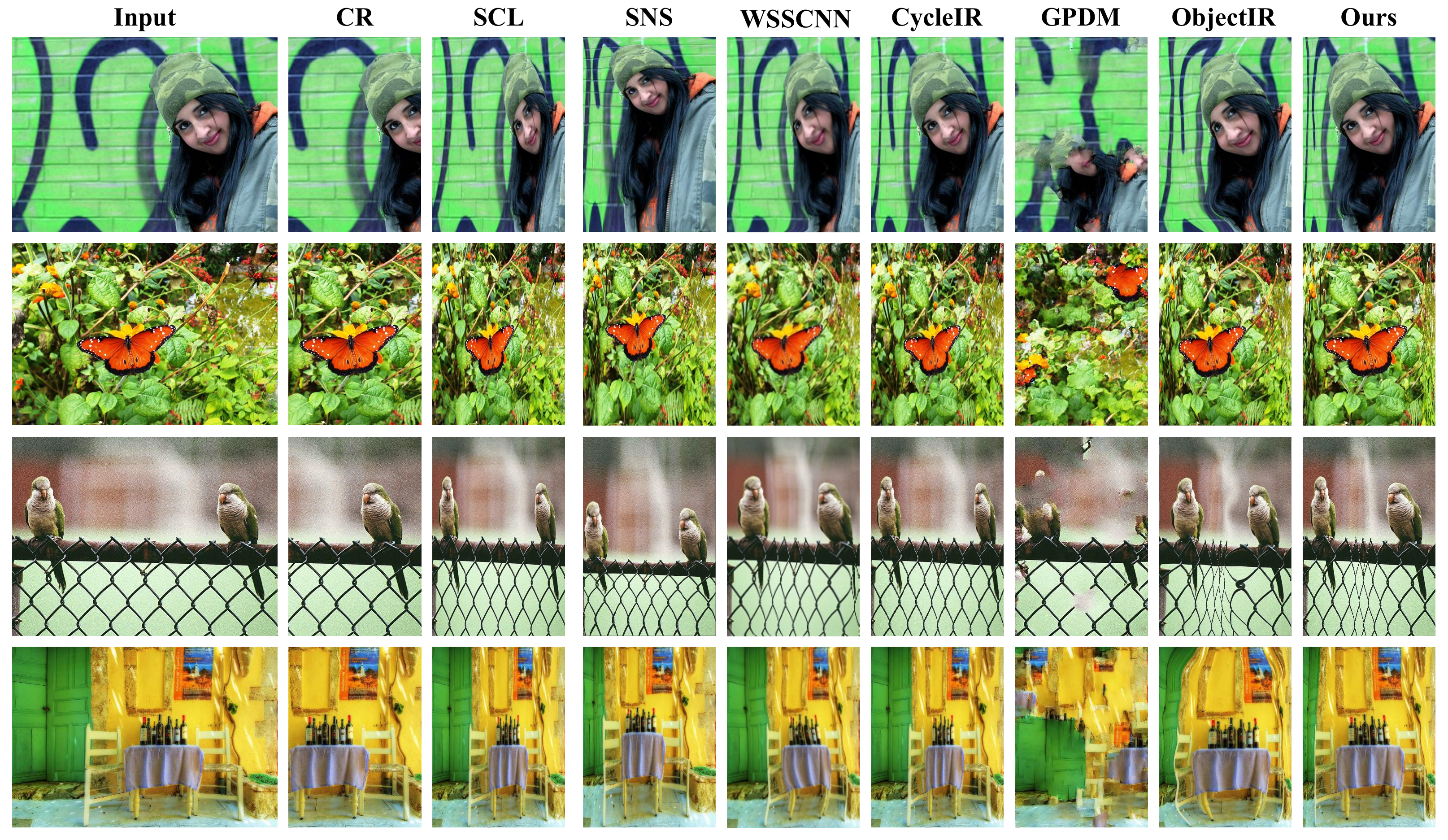}
	\caption{More visual comparisons on the RetargetMe dataset~\cite{rubinstein2010comparative}.
    All results are generated scaling the width to 0.5$\times$ its original value.
    From left to right are the original input, CR (manual cropping), SCL (simple scaling), SNS~\cite{wang2008optimized}, WSSCNN~\cite{cho2017weakly}, CycleIR~\cite{tan2019cycle}, GPDM~\cite{elnekave2022generating}, ObjectIR~\cite{liao2026object} and our method.}
    \label{fig:allMethods}
\end{figure*}

\begin{figure}[!t]
	\centering
	\includegraphics[width=\linewidth]{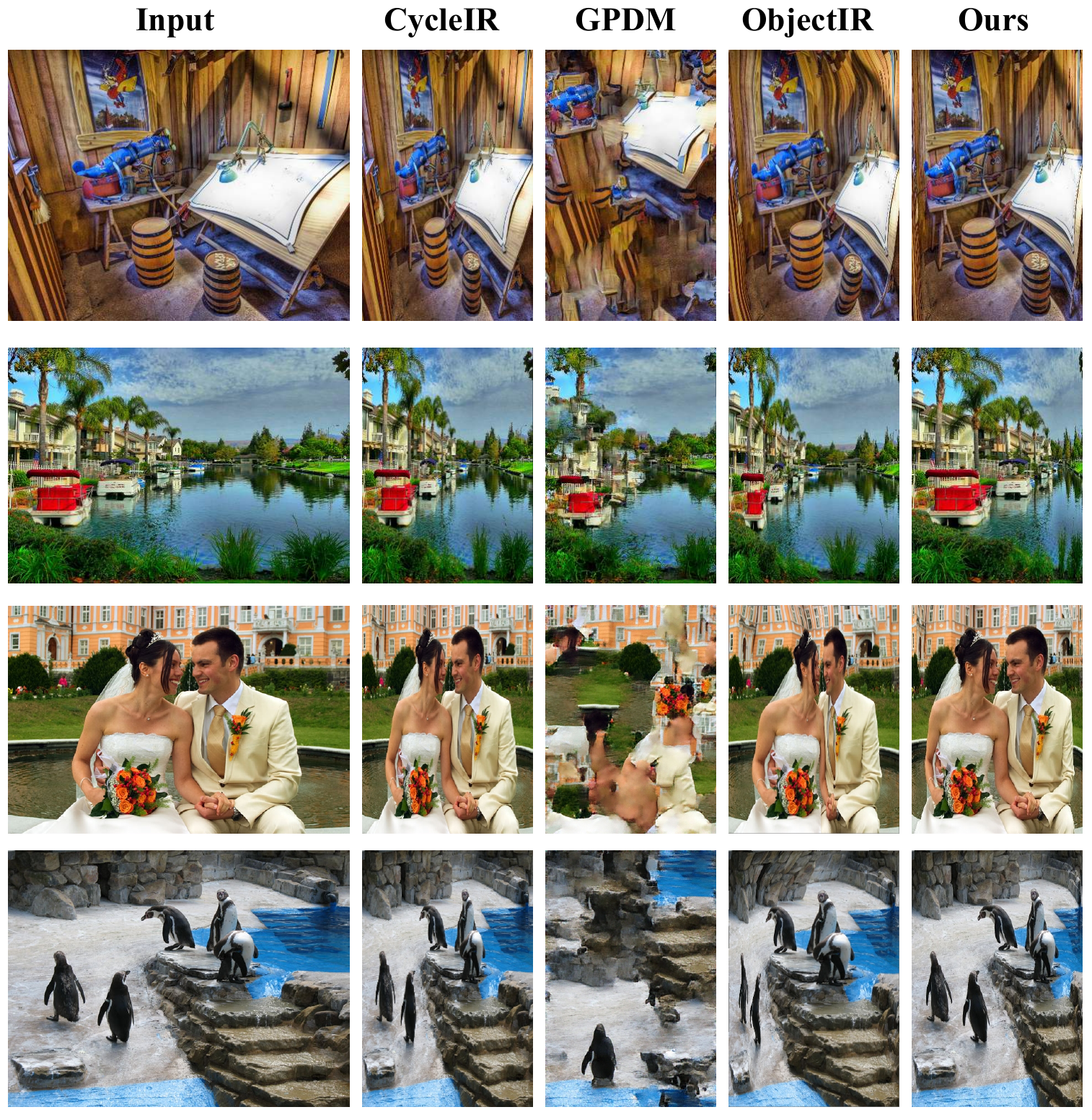}
	\caption{\ZY{Comparison on challenging examples with complex backgrounds (1--3 lines) and dispersed foreground regions (the last line)}.}
    \label{fig:challenging}
\end{figure}

\subsection{Qualitative Comparison}

\ZY{We first conduct a qualitative comparison across different scales, methods, and scenarios.
As shown in Fig.~\ref{differentScale}, we first compare our method with the SOTA retargeting method~\cite{liao2026object} across different scales.
Compared with this state-of-the-art method, our approach better preserves the scale and geometric structure of salient foreground objects at all tested scales, while simultaneously enabling reasonable deformation of background regions.
}

\ZY{
We further conduct a comprehensive comparison with representative methods, including traditional~\cite{wolf2007non,wang2008optimized} and learning-based~\cite{cho2017weakly,tan2019cycle,elnekave2022generating,liao2026object} methods.
As shown in Fig.~\ref{fig:allMethods}, it is easy to see that cropping (CR) can well preserve salient object but lose much contents, while simple scaling (SCL) severely distorted salient regions.
Other traditional content-aware warping methods can preserve the scale and structure of prominent content to a certain extent, yet their visual performance is slightly inferior to ours.
\ZZ{Most learning-based methods~\cite{cho2017weakly,tan2019cycle,liao2026object} learn content-aware image warping from data and are more efficient and robust than traditional methods, but they still face challenges such as salient-object and structural distortion. For learning-based generative methods such as GPDM~\cite{elnekave2022generating}, content fidelity remains a common drawback, and generated outputs may fail to maintain structurally coherent and semantically reasonable layouts.}
% In terms of the learning-based methods, most of them~\cite{cho2017weakly,tan2019cycle,liao2026object} learns to warping images from data in a content-aware manner. Although more efficient and robust than traditional methods,
% these learning-based solutions still faces challenges, such as salient object and structure distortion.
% For learning-based generative methods such as~\cite{elnekave2022generating}, content fidelity remains a common drawback. The generated outputs frequently fail to maintain structurally coherent and semantically reasonable layouts.
}

\ZY{
We finally evaluate the performance of our method in challenging examples.
% , where the salient regions exhibit complex, dispersed spatial distributions.
As illustrated in Fig.~\ref{fig:challenging}, for images with intricate background structures and widely scattered salient regions, ObjectIR~\cite{liao2026object} only imposes constraints on the bounding boxes of foreground salient regions while leaving background deformation unregulated, which results in unsatisfying retargeting on complex scenes.
Consequently, its outputs frequently suffer from distorted background textures and failure to preserve critical foreground information.
In contrast, our method introduces pixel-wise constraints on salient regions together with global structural regularization, yielding visually more favorable retargeting outputs.
}

\begin{figure*}[!t]
	\centering
	\includegraphics[width=\linewidth]{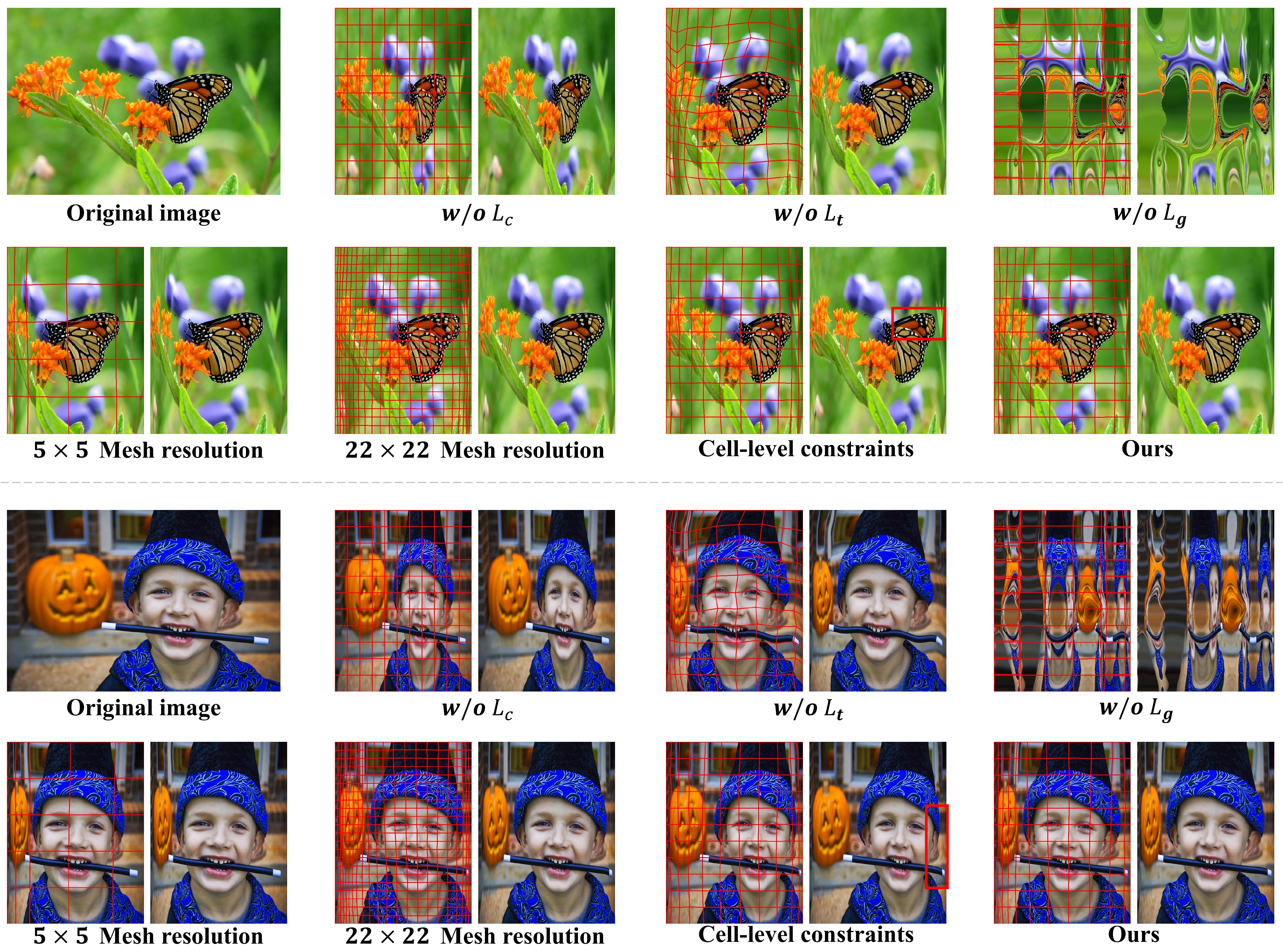}
	\caption{\ZY{Visual comparisons of ablation analysis on loss functions, mesh resolution, as well as pixel-level and cell-level constraints.}}
	\label{fig:ablation}
\end{figure*}

\subsection{Ablation Study}
\ZY{We conduct ablation studies on the ObjectIR~\cite{liao2026object} and RetargetMe~\cite{rubinstein2010comparative} datasets, to verify the effectiveness of our method, as shown in Table~\ref{tab:ablation} and Fig.~\ref{fig:ablation}.
}

\textbf{Loss functions.}
We first ablate the three losses to evaluate their individual contributions.
As illustrated in Rows 1–3 of Table~\ref{tab:ablation},
removing any loss term degrades the results  performance to different extents.
The visual results in Fig.~\ref{fig:ablation} further illustrate the role of each loss term.
Without $\mathcal{L}_c$, the model fails to sufficiently preserve salient content and tends to degenerate into nearly uniform scaling.

\textbf{Mesh resolution.}
We further explore the influence of mesh resolution via ablation experiments. Three typical mesh sizes are adopted: \(5 \times 5\), \(11 \times 11\), and \(22 \times 22\). As listed in the last 3 \ZZ{rows} of Table~\ref{tab:ablation}, the \(5 \times 5\) mesh yields inferior retargeting performance \ZZ{because its coarse deformation grid cannot capture fine structures.} 
In contrast, the two higher resolutions support delicate pixel warping and achieve visually superior results. Since the performance gap between \(11 \times 11\) and \(22 \times 22\) is marginal, we select \(11 \times 11\) as our final mesh resolution to strike a balance between model performance and computational efficiency.

\textbf{Pixel \& Cell-level constraints.}
We finally perform ablation analysis to investigate the impacts of mesh-level and pixel-level constraints imposed on visually salient regions.
Compared with the mesh-level constraint, as listed in Rows 5–6 of Table~\ref{tab:ablation}, the fine-grained pixel-level constraint makes full use of dense saliency information, leading to more favorable preservation of salient regions.
Results in Fig.~\ref{fig:ablation} further clarify the effectiveness of the pixel-level constraints.

% \begin{table}
% \centering
% \caption{User Study on the RetargetMe Dataset~\cite{rubinstein2010comparative}. 25 participants were invited to score 30 image groups randomly sampled from the RetargetMe dataset and generated by representative retargeting algorithms. The mean score and variance for each method are reported in the table.
% }
% \label{tab:user_study}
% \begin{tabular}{lcccc}
% \toprule
% Metric & Ours & Prune~\cite{shen2024prune} & SC~\cite{avidan2007seam} & ObjectIR~\cite{liao2026object} \\
% \midrule
% Avg. & \textbf{3.1040} & 2.7560 & 2.0240 & 2.3760 \\
% Var. & \textbf{1.0811} & 1.6383 & 1.2617 & 1.0998 \\
% \bottomrule
% \end{tabular}
% \end{table}

\begin{table*}
\centering
\caption{
\ZY{Ablation studies on the ObjectIR~\cite{liao2026object} and RetargetMe~\cite{rubinstein2010comparative} datasets.
The gray row corresponds to the results of our model, while the boldfaced and underlined entries indicate the best and second-best results, respectively.}
% Lower distortion error is better.
}
\label{tab:ablation}
\setlength{\tabcolsep}{9.0pt}
\renewcommand{\arraystretch}{1.15}
\begin{tabular}{c ccc cc ccc cc}
\toprule
\multirow{2}{*}{{ID}}
& \multicolumn{3}{c}{{Loss function}}
& \multicolumn{2}{c}{{Projection}}
& \multicolumn{3}{c}{{Mesh resolution}}
& \multicolumn{2}{c}{{Dataset}} \\
\cmidrule{2-11}
& $\mathcal{L}_{c}$
& $\mathcal{L}_{t}$
& $\mathcal{L}_{g}$
& Cell
& Pixel
& $5\times5$
& $11\times11$
& $22\times22$
& ObjectIR~\cite{liao2026object} 
& RetargetMe~\cite{rubinstein2010comparative}  \\
\midrule

1 
& 
& \checkmark
& \checkmark
& 
& \checkmark
& 
& \checkmark
& 
& 0.6391
& 0.7487 \\

2 
& \checkmark
& 
& \checkmark
& 
& \checkmark
& 
& \checkmark
& 
& 0.4032
& 0.5493 \\

3 
& \checkmark
& \checkmark
& 
& 
& \checkmark
& 
& \checkmark
& 
& 0.9876
& 1.0000 \\

4 
& \checkmark
& \checkmark
& \checkmark
& \checkmark
& 
& 
& \checkmark
& 
& 0.3096
& 0.4568 \\

\rowcolor{gray!20}
5 
& \checkmark
& \checkmark
& \checkmark
& 
& \checkmark
& 
& \checkmark
& 
& \underline{0.2917}
& \textbf{0.4420} \\

6 
& \checkmark
& \checkmark
& \checkmark
& 
& \checkmark
& \checkmark
& 
& 
& 0.3472
& 0.4771 \\

7 
& \checkmark
& \checkmark
& \checkmark
& 
& \checkmark
& 
& 
& \checkmark
& \textbf{0.2911}
& \underline{0.4625} \\

\bottomrule
\end{tabular}
\end{table*}

\begin{table}
\centering
\caption{User Study on the RetargetMe Dataset~\cite{rubinstein2010comparative}. 25 participants were invited to score 30 image groups randomly sampled from the RetargetMe dataset and generated by representative retargeting algorithms~\cite{shen2024prune,avidan2007seam,liao2026object}. The mean score and variance for each method are reported in the table.
}
\label{tab:user_study}
\begin{tabular}{lcccc}
\toprule
Metric & Ours & Prune\cite{shen2024prune} & SC\cite{avidan2007seam} & ObjectIR\cite{liao2026object} \\
\midrule
Avg. & \textbf{3.1040} & 2.7560 & 2.0240 & 2.3760 \\
Var. & \textbf{1.0811} & 1.6383 & 1.2617 & 1.0998 \\
\bottomrule
\end{tabular}
\end{table}

% \begin{table*}[t] 
% \centering \caption{ Ablation study of the proposed method. The gray row corresponds to our method with default parameter settings. } 
% \setlength{\tabcolsep}{6.0pt} 
% \renewcommand{\arraystretch}{1.15} 
% \begin{tabular}{c ccc ccc cc cc} 
% \toprule \multirow{2}{*}{ID} & \multicolumn{3}{c}{Loss function} & \multicolumn{3}{c}{Mesh resolution} & \multicolumn{2}{c}{Projection} & \multicolumn{2}{c}{Dataset} \\ 
% \cmidrule(lr){2-11}  & $\mathcal{L}_{c}$ & $\mathcal{L}_{t}$ & $\mathcal{L}_{g}$ & $5\times5$ & $11\times11$ & $22\times22$ & Cell & Pixel & Testing & RetargetMe \\
%  \midrule 1 & \checkmark & & & & \checkmark & & & \checkmark & 0.9975 & 1.0000 \\
% 2 & \checkmark & \checkmark & & & \checkmark & & & \checkmark & 0.9876 & 1.0000 \\
% 3 & \checkmark & & \checkmark & & \checkmark & & & \checkmark & 0.4032 & 0.5493 \\ 
% \rowcolor{gray!25} 4 & \checkmark & \checkmark & \checkmark & & \checkmark & & & \checkmark & 0.2917 & 0.4420 \\ 
% \midrule 5 & \checkmark & \checkmark & \checkmark & \checkmark & & & & \checkmark & 0.3472 & 0.4771 \\ 
% 6 & \checkmark & \checkmark & \checkmark & & & \checkmark & & \checkmark & 0.2911 & 0.4625 \\ 
% \midrule 7 & \checkmark & \checkmark & \checkmark & & \checkmark & & \checkmark & & 0.3096 & 0.4568 \\ 
% \bottomrule 
% \end{tabular} 
% \label{tab:ablation} 
% \end{table*}

\subsection{User Study}
We conduct a user study to quantitatively evaluate human preferences over retargeted outputs generated by different algorithms. Specifically, we take the dataset from~\cite{rubinstein2010comparative} as input images and \LT{generate} scaled retargeting results via several representative methods, including SC~\cite{avidan2007seam}, ObjectIR~\cite{liao2026object}, Prune~\cite{shen2024prune}, and our proposed approach. 
\LT{Each sample is randomly retargeted at two scales, 0.5$\times$ and 1.75$\times$ the original width, to evaluate the performance under both reduction and enlargement settings.}
We recruit 25 undergraduate participants, \LT{including 15 majoring in engineering and 10 from liberal arts and art-related majors, to independently evaluate randomly selected 30 groups of retargeted images.}
To eliminate subjective bias and guarantee fair evaluation, all participants \LT{are blinded to the source methods, and the order of results is randomly shuffled for each trial.}
% Meanwhile, all test images are randomly scaled up and down to ensure comprehensive evaluation.
Participants are asked to assign a score ranging from 1 to 5 (1 denotes the worst visual quality while 5 denotes the best) from three perspectives: structural integrity, visual naturalness and aesthetics, as well as the severity of geometric distortions and residual artifacts.
The results in Table.~\ref{tab:user_study} show that our method is preferred by most users. 
We further conduct a two-tailed paired t-test \LT{between} ObjectIR~\cite{liao2026object} and our methods, and the resulting p-value is far \LT{below} 0.05, which indicates a statistically significant \LT{improvement in visual quality over the SOTA retargeting baseline~\cite{liao2026object}}.

% \subsection{Discussion}

\enlargethispage{\baselineskip}
\section{Conclusion}
This paper proposes \textit{AttenIR}, an unsupervised image retargeting approach that requires no manual labels and is built upon the alignment between dense visual attention and mesh warping.
We model mesh deformation as a regression task constrained by multiple loss functions that preserve salient content, local shapes, and global structures without manual feature design or labeled data. Extensive ablation studies verify the effectiveness of the proposed pixel-level and mesh-level constraints, as well as the selected mesh resolution. Experimental results show that our method outperforms existing learning-based competitors in structural integrity and content fidelity.

\ZZ{In future work, we will extend our \LT{\textit{AttenIR}} framework to video retargeting and dynamic real-time scene adaptation. We also plan to incorporate aesthetic perception as an explicit evaluation and regularization criterion for image retargeting.}

% Place remaining floats before the references.
\FloatBarrier
{\fontsize{8.5}{10}\selectfont
\bibliographystyle{cvm}
\bibliography{references}

@inproceedings{wolf2007non,
  title={Non-homogeneous content-driven video-retargeting},
  author={Wolf, Lior and Guttmann, Moshe and Cohen-Or, Daniel},
  booktitle={2007 IEEE 11th international conference on computer vision},
  pages={1--6},
  year={2007},
  organization={IEEE}
}

@article{ChengPAMI,
  author = {Ming-Ming Cheng and Niloy J. Mitra and Xiaolei Huang and Philip H. S. Torr and Shi-Min Hu},
  title = {Global Contrast based Salient Region Detection},
  year  = {2015},
  journal= {IEEE TPAMI},
  volume={37}, 
  number={3}, 
  pages={569--582}, 
  doi = {10.1109/TPAMI.2014.2345401},
}

@article{SalObjBenchmark,
  author = {Ali Borji and Ming-Ming Cheng and Huaizu Jiang and Jia Li},
  title = {Salient Object Detection: A Benchmark},
  journal = {IEEE TIP},
  year={2015},
  volume={24},
  number={12},
  pages={5706-5722},
  doi={10.1109/TIP.2015.2487833}, 
}

@article{tan2019cycle,
  title={Cycle-IR: Deep cyclic image retargeting},
  author={Tan, Weimin and Yan, Bo and Lin, Chuming and Niu, Xuejing},
  journal={IEEE transactions on multimedia},
  volume={22},
  number={7},
  pages={1730--1743},
  year={2019},
  publisher={IEEE}
}

@article{liao2026object,
  title={Object-IR: Leveraging object consistency and mesh deformation for self-supervised image retargeting},
  author={Liao, Tianli and Wang, Ran and Zhang, Siqing and Li, Lei and Liu, Guangen and Zhao, Chenyang and Cao, Heling and Li, Peng},
  journal={Pattern Recognition},
  volume={172},
  pages={112651},
  year={2026}
}

@article{kim_quad_2019,
  title = {A Quad Edge-Based Grid Encoding Model for Content-Aware Image Retargeting},
  author = {Kim, Yoonhyung and Eun, Hyunjun and Jung, Chanho and Kim, Changick},
  year = 2019,
  month = dec,
  journal = {IEEE Transactions on Visualization and Computer Graphics},
  volume = {25},
  number = {12},
  pages = {3202--3215},
  issn = {1077-2626},
  doi = {10.1109/TVCG.2018.2866106}
}

@incollection{wang2008optimized,
  title={Optimized scale-and-stretch for image resizing},
  author={Wang, Yu-Shuen and Tai, Chiew-Lan and Sorkine, Olga and Lee, Tong-Yee},
  booktitle={ACM SIGGRAPH Asia 2008 papers},
  pages={1--8},
  year={2008},
  publisher={ACM}
}

@inproceedings{shen2024prune,
  title={Prune and Repaint: Content-Aware Image Retargeting for Any Ratio},
  author={Shen, Feihong and Li, Chao and Geng, Yifeng and Deng, Yongjian and Chen, Hao},
  booktitle={Advances in Neural Information Processing Systems},
  volume={37},
  pages={75177--75198},
  year={2024},
  doi={10.52202/079017-2393}
}

@inproceedings{gao2024mdsam,
  title={Multi-Scale and Detail-Enhanced Segment Anything Model for Salient Object Detection},
  author={Gao, Shixuan and Zhang, Pingping and Yan, Tianyu and Lu, Huchuan},
  booktitle={Proceedings of the 32nd ACM International Conference on Multimedia},
  pages={9894--9903},
  year={2024},
  doi={10.1145/3664647.3680650}
}

@article{avidan2007seam,
  title={Seam carving for content-aware image resizing},
  author={Avidan, Shai and Shamir, Ariel},
  journal={ACM Transactions on Graphics},
  volume={26},
  number={3},
  pages={10},
  year={2007},
  doi={10.1145/1276377.1276390}
}

@article{dong2009optimized,
  title={Optimized image resizing using seam carving and scaling},
  author={Dong, Weiming and Zhou, Ning and Paul, Jean-Claude and Zhang, Xiaopeng},
  journal={ACM Transactions on Graphics},
  volume={28},
  number={5},
  pages={1--10},
  year={2009},
  doi={10.1145/1618452.1618471}
}

@article{zhang2009shape,
  title={A shape-preserving approach to image resizing},
  author={Zhang, Guo-Xin and Cheng, Ming-Ming and Hu, Shi-Min and Martin, Ralph R.},
  journal={Computer Graphics Forum},
  volume={28},
  number={7},
  pages={1897--1906},
  year={2009},
  doi={10.1111/j.1467-8659.2009.01568.x}
}

@article{guo2009mesh,
  title={Image retargeting using mesh parametrization},
  author={Guo, Yanwen and Liu, Feng and Shi, Jian and Zhou, Zhihua and Gleicher, Michael},
  journal={IEEE Transactions on Multimedia},
  volume={11},
  number={5},
  pages={856--867},
  year={2009},
  doi={10.1109/TMM.2009.2021781}
}

@inproceedings{chang2012line,
  title={A line-structure-preserving approach to image resizing},
  author={Chang, Che-Han and Chuang, Yung-Yu},
  booktitle={IEEE Conference on Computer Vision and Pattern Recognition, {CVPR}},
  pages={1075--1082},
  year={2012},
  organization={IEEE},
  doi={10.1109/CVPR.2012.6247785}
}

@article{ZHANG2026104615,
title = {Revisiting unsupervised image stitching via efficient boundary rectification},
journal = {Computers \& Graphics},
volume = {137},
pages = {104615},
year = {2026},
issn = {0097-8493},
doi = {https://doi.org/10.1016/j.cag.2026.104615},
author = {Yun Zhang and Jialing Yang and Ruiyang Liang and Yao Xu and Lang Nie and Jinyu Xu and Fang-Lue Zhang and Xinyuan Zheng},

}

@inproceedings{DBLP:conf/cvpr/NieLLL022,
  author       = {Lang Nie and
                  Chunyu Lin and
                  Kang Liao and
                  Shuaicheng Liu and
                  Yao Zhao},
  title        = {Deep Rectangling for Image Stitching: {A} Learning Baseline},
  booktitle    = {{IEEE/CVF} Conference on Computer Vision and Pattern Recognition (CVPR)},
  pages        = {5730--5738},
  publisher    = {{IEEE}},
  year         = {2022},
}

@inproceedings{TPS_Transformation,
author = {Jaderberg, Max and Simonyan, Karen and Zisserman, Andrew and Kavukcuoglu, Koray},
title = {Spatial transformer networks},
year = {2015},
publisher = {MIT Press},
address = {Cambridge, MA, USA},
booktitle = {Proceedings of the 29th International Conference on Neural Information Processing Systems - Volume 2},
pages = {2017–2025},
numpages = {9},
location = {Montreal, Canada},
series = {NIPS'15}
}

@inproceedings{DBLP:conf/cvpr/HeZRS16,
  author       = {Kaiming He and
                  Xiangyu Zhang and
                  Shaoqing Ren and
                  Jian Sun},
  title        = {Deep Residual Learning for Image Recognition},
  booktitle={IEEE Conference on Computer Vision and Pattern Recognition, {CVPR}},
  pages        = {770--778},
  organization={IEEE},
  year         = {2016},
  doi          = {10.1109/CVPR.2016.90},
}

@article{panozzo2012robust,
  title={Robust image retargeting via axis-aligned deformation},
  author={Panozzo, Daniele and Weber, Ofir and Sorkine, Olga},
  journal={Computer Graphics Forum},
  volume={31},
  number={2pt1},
  pages={229--236},
  year={2012},
  doi={10.1111/j.1467-8659.2012.03001.x}
}

@article{niu2020semantic,
  title={Semantic segmentation guided pixel fusion for image retargeting},
  author={Niu, Xuejing and Bare, Bahetiyaer and Tan, Weimin},
  journal={IEEE Transactions on Multimedia},
  volume={22},
  number={3},
  pages={676--687},
  year={2020},
  doi={10.1109/TMM.2019.2932566}
}

@article{cao2023polygonal,
  title={Polygonal finite element-based content-aware image warping},
  author={Cao, Jun and Zhang, Hong and Liu, Ligang and Yang, Xiaosong and Liu, Shiguang and Xu, Ying and Chen, Jun},
  journal={Computational Visual Media},
  volume={9},
  number={2},
  pages={325--338},
  year={2023},
  doi={10.1007/s41095-022-0284-8}
}

@inproceedings{cho2017weakly,
  title={Weakly- and self-supervised learning for content-aware deep image retargeting},
  author={Cho, Donghyeon and Park, Jinsun and Oh, Tae-Hyun and Tai, Yu-Wing and Kweon, In So},
  booktitle={Proceedings of the IEEE International Conference on Computer Vision},
  pages={4568--4577},
  year={2017},
  doi={10.1109/ICCV.2017.488}
}

@inproceedings{lin2019deepir,
  title={DeepIR: A deep semantics driven framework for image retargeting},
  author={Lin, Jianxin and Zhou, Tiankuang and Chen, Zhibo},
  booktitle={2019 IEEE International Conference on Multimedia and Expo Workshops},
  pages={54--59},
  year={2019},
  organization={IEEE},
  doi={10.1109/ICMEW.2019.0-111}
}

@article{danon2021deepfeatures,
  title={Image resizing by reconstruction from deep features},
  author={Danon, Dolev and Arar, Moab and Cohen-Or, Daniel and Shamir, Ariel},
  journal={Computational Visual Media},
  volume={7},
  number={4},
  pages={453--466},
  year={2021},
  doi={10.1007/s41095-021-0223-6}
}

@article{le2024retargeting,
  title={Retargeting video with an end-to-end framework},
  author={Le, Thi Ngoc Hanh and Huang, Hui Guang and Chen, Yi Ru and Lee, Tong Yee},
  journal={IEEE Transactions on Visualization and Computer Graphics},
  volume={30},
  number={9},
  pages={6164--6176},
  year={2024},
  doi={10.1109/TVCG.2023.3327825}
}

@inproceedings{xu2026halo,
  title={HALO: Human-Aligned End-to-end Image Retargeting with Layered Transformations},
  author={Xu, Yiran and Xie, Siqi and Li, Zhuofang and Shadmany, Harris and Li, Yinxiao and Sbaiz, Luciano and Wang, Miaosen and Ke, Junjie and Lezama, Jos{\'e} and Qi, Hang and others},
  booktitle    = {{IEEE/CVF} Conference on Computer Vision and Pattern Recognition (CVPR)},
  publisher    = {{IEEE}},
  pages={2733--2742},
  year={2026}
}

@inproceedings{dong2026airdr,
  title={AIR-DR: Adaptive Image Retargeting with Instance Relocation and Dual-Guidance Repainting},
  author={Dong, Zhitong and Li, Chao and Deng, Yongjian and Chen, Hao},
  booktitle={Proceedings of the AAAI Conference on Artificial Intelligence},
  pages={3668--3676},
  year={2026}
}

@article{ma2012quality,
  title={Image retargeting quality assessment: A study of subjective scores and objective metrics},
  author={Ma, Lin and Lin, Weisi and Deng, Chenwei and Ngan, King Ngi},
  journal={IEEE Journal of Selected Topics in Signal Processing},
  volume={6},
  number={6},
  pages={626--639},
  year={2012},
  doi={10.1109/JSTSP.2012.2211996}
}

@article{guo2023irqa,
  title={Image retargeting quality assessment: A survey},
  author={Guo, Yingchun and Wang, Dan and Yan, Gang and Zhu, Ye},
  journal={Journal of Intelligent \& Fuzzy Systems},
  volume={44},
  number={2},
  pages={1921--1942},
  year={2023},
  doi={10.3233/JIFS-220456}
}

@article{fan2024review,
  title={A comprehensive review of image retargeting},
  author={Fan, Xiaoting and Zhang, Zhong and Sun, Long and Xiao, Baihua and Li, Ping},
  journal={Neurocomputing},
  volume={579},
  pages={127416},
  year={2024},
  doi={10.1016/j.neucom.2024.127416}
}

@article{rubinstein2008improved,
  title={Improved seam carving for video retargeting},
  author={Rubinstein, Michael and Shamir, Ariel and Avidan, Shai},
  journal={ACM Transactions on Graphics},
  volume={27},
  number={3},
  pages={16:1--16:9},
  year={2008},
}

@article{rubinstein2009multi,
  title={Multi-operator media retargeting},
  author={Rubinstein, Michael and Shamir, Ariel and Avidan, Shai},
  journal={ACM Transactions on Graphics},
  volume={28},
  number={3},
  pages={23:1--23:11},
  year={2009},
}

@inproceedings{simakov2008summarizing,
  title={Summarizing visual data using bidirectional similarity},
  author={Simakov, Denis and Caspi, Yaron and Shechtman, Eli and Irani, Michal},
  booktitle={Proceedings of the IEEE Conference on Computer Vision and Pattern Recognition},
  pages={1--8},
  year={2008},
  doi={10.1109/CVPR.2008.4587842}
}

@inproceedings{pritch2009shift,
  title={Shift-map image editing},
  author={Pritch, Yael and Kav-Venaki, Eitam and Peleg, Shmuel},
  booktitle={Proceedings of the IEEE International Conference on Computer Vision},
  pages={151--158},
  year={2009},
  doi={10.1109/ICCV.2009.5459159}
}

@article{rubinstein2010comparative,
  title={A comparative study of image retargeting},
  author={Rubinstein, Michael and Gutierrez, Diego and Sorkine, Olga and Shamir, Ariel},
  journal={ACM Transactions on Graphics},
  volume={29},
  number={6},
  pages={160:1--160:10},
  year={2010},
  doi={10.1145/1882261.1866186}
}

@article{liu2011quality,
  title={Image retargeting quality assessment},
  author={Liu, Yong-Jin and Luo, Xi and Xuan, Yu-Ming and Chen, Wen-Feng and Fu, Xiao-Lan},
  journal={Computer Graphics Forum},
  volume={30},
  number={2},
  pages={583--592},
  year={2011},
  doi={10.1111/j.1467-8659.2011.01881.x}
}

@inproceedings{elsner2024deformation,
  title={Retargeting visual data with deformation fields},
  author={Elsner, Tim and Berger, Julia and Wu, Tong and Czech, Victor and Gao, Lin and Kobbelt, Leif},
  booktitle={Computer Vision -- ECCV 2024},
  pages={271--288},
  year={2024},
  publisher={Springer},
}

@inproceedings{song2018photo,
  title={Photo squarization by deep multi-operator retargeting},
  author={Song, Yu and Tang, Fan and Dong, Weiming and Zhang, Xiaopeng and Deussen, Oliver and Lee, Tong-Yee},
  booktitle={Proceedings of the 26th ACM International Conference on Multimedia},
  pages={1047--1055},
  year={2018},
}

@inproceedings{lin2014microsoft,
  title={Microsoft coco: Common objects in context},
  author={Lin, Tsung-Yi and Maire, Michael and Belongie, Serge and Hays, James and Perona, Pietro and Ramanan, Deva and Doll{\'a}r, Piotr and Zitnick, C Lawrence},
  booktitle={European conference on computer vision},
  pages={740--755},
  year={2014},
  organization={Springer}
}

@inproceedings{elnekave2022generating,
  title={Generating natural images with direct patch distributions matching},
  author={Elnekave, Ariel and Weiss, Yair},
  booktitle={European Conference on Computer Vision},
  pages={544--560},
  year={2022},
  organization={Springer}
}

@article{ding2020image,
  title={Image quality assessment: Unifying structure and texture similarity},
  author={Ding, Keyan and Ma, Kede and Wang, Shiqi and Simoncelli, Eero P},
  journal={IEEE transactions on pattern analysis and machine intelligence},
  volume={44},
  number={5},
  pages={2567--2581},
  year={2020},
  publisher={IEEE}
}

@inproceedings{ke2021musiq,
  title={Musiq: Multi-scale image quality transformer},
  author={Ke, Junjie and Wang, Qifei and Wang, Yilin and Milanfar, Peyman and Yang, Feng},
  booktitle={Proceedings of the IEEE/CVF international conference on computer vision},
  pages={5148--5157},
  year={2021}
}
}
\end{document}